\pdfoutput=1

\documentclass[11pt]{article}

\usepackage[final]{acl}

\usepackage{times}
\usepackage{latexsym}

\usepackage[T1]{fontenc}

\usepackage[utf8]{inputenc}

\usepackage{microtype}

\usepackage{inconsolata}

\usepackage{natbib}
\usepackage{booktabs}
\usepackage{hyperref}
\usepackage[printonlyused,withpage]{acronym}
\usepackage{graphicx}
\usepackage{caption}
\usepackage{subcaption}
\usepackage[acronym]{glossaries}
\glsdisablehyper
\usepackage{siunitx}
\usepackage{amssymb}
\usepackage{wrapfig}
\usepackage{tabularx,ragged2e,booktabs}
\usepackage[flushleft]{threeparttable}
\usepackage{tablefootnote}
\usepackage{array}
\usepackage{multirow}
\usepackage{makecell}
\newcolumntype{T}{>{\footnotesize}l} \newacronym{LM}{LM}{Language Model}
\newacronym{LLM}{LLM}{Large Language Model}
\newacronym{QA}{QA}{Question Answering}
\newacronym{BPE}{BPE}{Byte-Pair Encoding}
\newacronym{BBPE}{BBPE}{Byte-Level BPE}
\newacronym{SP}{SP}{SentencePiece}
\newacronym{HF}{HF}{Hugging Face}
\newacronym{NLP}{NLP}{Natural Language Processing}

\title{Investigating Multilingual Instruction-Tuning: Do Polyglot Models Demand for Multilingual Instructions?}

 \author{Alexander Arno Weber\textsuperscript{1,2} Klaudia Thellmann\textsuperscript{3} Jan Ebert\textsuperscript{4} Nicolas Flores-Herr\textsuperscript{1} \AND Jens Lehmann$^{3}$ Michael Fromm\textsuperscript{1,2} Mehdi Ali\textsuperscript{1,2} \\   \\
  \textsuperscript{1}Lamarr Institute \textsuperscript{2}Fraunhofer IAIS \textsuperscript{3}TU Dresden \textsuperscript{4}FZ Jülich
}

\begin{document}
\maketitle
\begin{abstract}

The adaption of multilingual pre-trained \glspl{LLM} into eloquent and helpful assistants is essential to facilitate their use across different language regions.
In that spirit, we are the first to conduct an extensive study of the performance of multilingual models instruction-tuned on different language compositions on \textit{parallel} \textit{instruction-tuning} benchmarks across a selection of the most spoken Indo-European languages. 
We systematically examine the effects of language and instruction dataset size on a mid-sized and a large, multilingual \glspl{LLM} by instruction-tuning them on parallel instruction-tuning datasets.
Our results demonstrate that instruction-tuning on parallel instead of monolingual corpora benefits cross-lingual instruction following capabilities by up to 
9.9\%.
Furthermore, we show that the \textit{Superficial Alignment Hypothesis} does not hold in general, as the investigated multilingual 7B parameter model presents a counter-example requiring large-scale instruction-tuning datasets.
Finally, we conduct a human annotation study to understand the alignment between human-based and GPT-4-based evaluation within multilingual chat scenarios.

\end{abstract}

\section{Introduction}
\glspl{LLM} have a significant impact on the daily work of many, as they 
are practical to use and assist in solving natural text problems ranging from creative writing to math problems~\cite{DBLP:conf/emnlp/DingCXQHL0Z23ultrachat}.
One of the primary reasons for their fast adoption as assistants is their facilitated usage by simply instructing the model to conduct a specific task.
The training of such an assistant involves multiple stages of model training~\cite{DBLP:conf/acl/MishraKBH22naturalinstruct}.
First, an extensive, compute-intensive pre-training over large document corpora is conducted where the model is typically trained to predict the next token in a sequence.
While pre-trained \glspl{LLM} already exhibit significant capabilities, there is a discrepancy between their pre-training objective and the users' need for an instruction-following model.
Therefore, in the second step, fine-tuning is performed where the model is trained to follow the users' instructions within a chat environment.
The second step is crucial for the model to solve complex, multi-turn user requests.

With the availability of strong open-source English-centric models~\citep{touvronLlama2OpenFoundation2023}, many English-centric instruction-tuning datasets emerged~\citep{wangSelfInstructAligningLanguage2023,longpreFlanCollectionDesigning2023,zhouLIMALessMore2023}.
While there are adoptions of monolingual English models for other languages~\citep{uhligPHOENIXOpenSourceLanguage2024}, the choice and training of tokenizers is critical for the fair representation of languages~\citep{petrovLanguageModelTokenizers2023,aliTokenizerChoiceLLM2023} within pre-trained models, as otherwise, underrepresented languages are not only more resource intensive regarding training and inference, but also the performance for those languages drop.
A fundamental problem is the availability of appropriate open-source, multilingual datasets and benchmarks for training and assessing instruction-tuned \glspl{LLM}.
Here, especially the lack of multilingual benchmarks targeting instruction-tuned models represents a major gap, as previous instruction-tuned multilingual models are only evaluated on zero- or few-shot, academic benchmarks targeting pre-trained \glspl{LLM}~\citep{muennighoffCrosslingualGeneralizationMultitask2023,holmstromMakingInstructionFinetuningSwedish2023}.
However, it is essential to evaluate the multilingual instruction-following capabilities of the model on instruction benchmarks to realistically assess the helpfulness of a model as a chat assistant.

To the best of our knowledge, there is no study investigating multilingual instruction-tuning focusing on multilingual user request performance where the base model has been pre-trained with a large amount of multilingual data where English is represented only by around 42\%.
We tackle this research gap by translating \textit{MT-Bench} into the parallel benchmark \textit{MT-Bench-X} and systematically investigate how the language and size of instruction datasets impact the instruction-tuning of pre-trained, mid-sized, and large multilingual \glspl{LLM} for the Germanic 
and Italo-Western language family, including English, German, French, Italian, and Spanish, on this novel benchmark dataset.

To answer the research question of whether multilingual models pre-trained with a focus on Indo-European languages require instruction-tuning in all target languages to show competitive instruction-following capabilities across target languages, we make the following contributions:

\begin{itemize}
    \item Creation of \textit{Lima-X}\footnote{\url{https://hf.co/datasets/lamarr-org/Lima-X}}, a high-quality, complex, parallel corpus comprising 1030 instructions for each English, German, French, Italian, and Spanish (Section~\ref{sec:instruction-tuning}).
    \item Creation of \textit{MT-Bench-X}\footnote{\url{https://github.com/Modalities/MT-Bench-X}}, a parallel, multilingual, human-curated evaluation dataset for evaluating instruction-tuned \glspl{LLM} (Section~\ref{sec:instruct-tuning-evaluation}).
    \item Study about the effect of i.) the dataset size, ii.) the dataset nature, i.e., human-curated or synthetic data, iii.) the dataset language composition, and iv.) the importance of the predominant language during pre-training on multilingual instruction for a medium-sized and large language model (Section~\ref{sec:instruct-tuning-results}).
    \item Correlation analysis of the agreement levels between human experts and GPT-4 on MT-Bench-X (Section~\ref{sec:pair_wise_human_eval}).
\end{itemize}

\section{Related Work}
This section provides an overview of instruction-tuning datasets and aspects important for their utilization.

\subsection{Instruction-tuning}

\paragraph{Template-based Datasets}
Several English-focused instruction-tuning datasets have been introduced 
to broaden the scope of tasks and response formats by incorporating diverse sets of instructions~\citep{iyerOPTIMLScalingLanguage2022, longpreFlanCollectionDesigning2023}.
Primarily, many of these datasets revolve around Natural Language Processing (NLP) benchmarks that are refined through the application of either single or multiple prompt templates for responses and requests~\citep{longpreFlanCollectionDesigning2023, muennighoffCrosslingualGeneralizationMultitask2023}.

\paragraph{Synthetic Datasets}
An alternative approach involves extending only requests of NLP benchmarks by templates but letting sophisticated instruction-tuned models predict responses \citep{zhangInstructionTuningLargeSurvey2023}.
Examples here are 
\textit{OASST}~\citep{kopfOpenAssistantConversationsDemocratizing2023} and \textit{LIMA}~\citep{zhouLIMALessMore2023}.
The latter introduces the \textit{Superficial Alignment Hypothesis}~\citep{kirstain-etal-2022-examples, zhouLIMALessMore2023}.
It states that only a few examples per task or instruction format are required to teach a \gls{LLM} the response style.
At the same time, most of the capabilities and knowledge are acquired during pre-training.
While gaining great performance advancements with instructional data ranked by user preferences~\citep{uhligPHOENIXOpenSourceLanguage2024,laiOkapiInstructiontunedLarge2023}, this study focuses on the role of language mixtures during instruction-tuning.

\paragraph{Multilingual Datasets}
\citet{muennighoffCrosslingualGeneralizationMultitask2023} translate classical NLP tasks
, which were refined by prompt templates to create the \textit{xP3mt} dataset.
The authors indicate that fine-tuning solely in English is adequate for a multilingual pre-trained \gls{LLM} to adapt and perform well across various tasks in other pre-trained languages.
However, these results were evaluated solely on downstream evaluation tasks for pre-trained \glspl{LLM} and not on evaluation schemes developed for evaluating instruction-tuned models. 
On the other hand, \citet{holmstromMakingInstructionFinetuningSwedish2023} translate and evaluate instruction-tuning datasets for Swedish and their results indicate translated instructions significantly improve zero-shot performance of models and strong foundation in the target language benefits model performance, which contradicts the findings of \citet{muennighoffCrosslingualGeneralizationMultitask2023}.
This discrepancy might be introduced by the lack of response diversity~\citep{liBactrianXMultilingualReplicable2023} of template-based datasets and their dissimilarity to human-created and generated responses as Table~\ref{tab:dataset_stats} of the preliminary study in the Appendix suggests.

Bactrian-X~\citep{liBactrianXMultilingualReplicable2023} is based on machine-translated user requests, whereas corresponding responses were generated by OpenAI's \texttt{gpt-3.5-turbo}.

\subsection{Evaluation}
\paragraph{Pre-training Benchmarks}
Most often multilingual benchmarks, such as XCOPA~\citep{ponti-etal-2020-xcopa} and XWinograd~\citep{tikhonov-ryabinin-2021-heads} are utilized~\citep{muennighoffCrosslingualGeneralizationMultitask2023, liBactrianXMultilingualReplicable2023, laiOkapiInstructiontunedLarge2023} for evaluating instruction-tuned models across languages.
While these benchmarks measure specific aspects of \textit{pre-trained} \glspl{LLM} by accuracy regarding a gold truth often only spanning only a few words, they fail to capture the complex diversity instruction responses may offer~\citep{zhengJudgingLLMasajudgeMTBench2023}. 

\paragraph{Instruction-tuning Benchmarks}
With \textit{MT-Bench}, \citet{zhengJudgingLLMasajudgeMTBench2023} developed an automatic benchmark that encompasses the diversity of instruction responses by utilizing \textit{LLMs-as-a-judge}. Despite the availability of recent alternatives~\citep{liuGEvalNLGEvaluation2023,ferronMEEPThisEngaging2023,yeFLASKFinegrainedLanguage2023a}, \textit{MT-Bench} was broadly adopted by the community\footnote{\url{https://paperswithcode.com/dataset/mt-bench}}.

\citet{singhAyaDatasetOpenAccess2024,ustunAyaModelInstruction}, introduces the \textit{Aya Evaluation Suite}, which covers multilingual open-ended generation evaluation across 101 languages. While their prompts are suited for the conversational setup, a key difference to \textit{MT-Bench-X} is that it only covers single turns.

The work of Li et al. focuses on low-resource language instruction-tuning on the English-centric Llama-2~\cite{touvronLlama2OpenFoundation2023} model. In contrast, in this work, we focus on the composition of language mixture datasets of high-resource European languages on multilingual models by design and underline the importance of thoroughly evaluated multilingual human-curated evaluation datasets.

\subsection{Demarcation}
While works exist addressing multilingual fine-tuning, our work differs from others in central aspects:
\begin{itemize}
    \item We conduct our instruction-tuning based on mid-sized (7B) and large (8x7B) pre-trained models that have been trained with a focus on Indo-European languages.
    \item We investigate whether an instruction tuning dataset needs to be represented in each language in parallel, has to be split across languages, or should be monolingual.
    \item We inspect the role of multilingual dataset quality by comparing models trained on human-curated and translated datasets with models trained on same-sized synthetically generated datasets.
    \item We evaluate our models on manually human-curated multilingual benchmarks that specifically assess the model's instruction-following capabilities.
\end{itemize}

\section{Multilingual Instruction-Tuning Data}
\label{sec:instruction-tuning}

To investigate the defined research questions, we require high-quality parallel instruction-tuning datasets of different sizes.
While there exist multilingual instruction datasets, the distribution of languages is highly skewed towards English or datasets contain shorter, less complex responses~\citep{muennighoffCrosslingualGeneralizationMultitask2023}, as a preliminary study shown in the Appendix~\ref{sec:dataset_study} reveals. 
An exception here is Bactrian-X~\citep{liBactrianXMultilingualReplicable2023}, a large-scale, parallel instruction-tuning dataset.
Therefore, we select Bactrian-X (Section~\ref{data:Bactrian_X}) as a large-scale multilingual instruction-tuning dataset and translate a smaller English-only instruction-tuning dataset \textit{LIMA}~\citep{zhouLIMALessMore2023} into our target languages (Section~\ref{data:Lima_X}).
For both datasets, we created different multilingual compositions (Section~\ref{data:language-mixtures})\footnote{The license details of our created datasets are described in Appendix~\ref{appendix:licenses}}.

\subsection{Bactrian-X}\label{data:Bactrian_X}
The large-scale instruction-tuning dataset \textit{Bactrian-X}~\citep{liBactrianXMultilingualReplicable2023} was created by translating the original English instructions and generating responses by ChatGPT (\texttt{gpt-3.5-turbo}).
We selected English, German, Italian, French and Spanish as target languages.
We keep 95\% (64K samples per language) of the data as the training dataset and use the remaining 5\% as the validation dataset (3K samples per language).
We additionally down-sample the dataset to the same size as \textit{LIMA}.

\subsection{Lima-X}\label{data:Lima_X}
Each sample in \textit{LIMA} is carefully curated, which is one benefit of its manageable size of 1030 samples.
Despite the creation of a validation set with high standards of curation by \citet{zhouLIMALessMore2023},
the authors did not publish their validation dataset.
Simply sampling the validation dataset from a training data split might remove samples providing important learning signals that are potentially not redundant within the remaining few samples.
We thus adapt the curation steps and create a novel validation dataset that is described in the Appendix~\ref{appendix:lima-val}.

As we focus on Indo-European languages in our study, we chose to utilize DeepL as a translator performing well in these languages \cite{yulianto2021google, jiaoChatGPTGoodTranslator2023, hidalgo2020google}.
We translate \textit{LIMA} and the novel validation dataset into German, French, Italian, and Spanish.
Before translating, we manually reviewed all 1030 training instances and marked the ones that could lead to problematic translations. 
The reasons here could be i.) mixed language usage in text, ii.) code snippets, where code comments should be translated into other languages, but control statements not,
iii.) samples that were written entirely in a language other than English, and iv.) cultural aspects of English that are not transferable to the target language.
We mark 66 such cases in total and investigate whether DeepL can handle those for German.
The \textit{LIMA} dataset has 12 entries of non-English language, e.g., Portuguese, Spanish, or German, which are not problematic for DeepL.
However, variable names in code snippets were partially translated.
Furthermore, jokes and poems are not directly translatable, which we see as a downside of the translation approach.

\subsection{Language Mixtures}\label{data:language-mixtures}
We compose multilingual variants of the translated monolingual datasets in our five target languages that make up \textit{Lima-X} and our language selection of \textit{Bactrian-X}.
This involves creating \texttt{ENDEFRITES}, which encompasses all five languages, effectively increasing the dataset size five-fold. In the same manner, we create \texttt{DEFRITES}, which does not contain the English part of the datasets and thus only quadruples the dataset size.
Additionally, we create a variant, maintaining the same semantics of the questions as in the monolingual original but distributed equally across the five languages within the dataset, resulting in the concrete datasets \texttt{ENDEFRITES-sampled} and \textit{DEFRITES-sampled}. 
For example, originating from the English dataset example ID \textit{100}, \texttt{ENDEFRITES-sampled} will contain either one of \textit{100\_X} by choice,  whereas \textit{X} stands for the fine-tuning language \textit{EN}, \textit{DE}, \textit{FR}, \textit{IT}, \textit{ES} and so that no ID is duplicated across languages and all languages are represented equally often. To analyze the role of the dataset nature, i.e., human-curated vs. synthetic datasets, we downsample the Bactrian-X dataset variants to contain as many examples as Lima-X, utilizing the same seed across languages, resulting in the dataset variants \textit{Bactrian-X-small}.

\section{Multilingual Benchmarks}
\label{sec:instruct-tuning-evaluation}

To evaluate the multilingual instruction-following capabilities of the models, a comprehensive multilingual benchmark for our target languages is indispensable.
Thus, we translated MT-Bench~\cite{zhengJudgingLLMasajudgeMTBench2023} (Section~\ref{bench:mt_bench}) into the five target languages, followed by human quality assurance for each language to create the multilingual MT-Bench variant MT-Bench-X (Section~\ref{sec:translation-mtbenchx}).
We employed MT-Bench-X to conduct a machine evaluation and a human evaluation (Section~\ref{sec:eval-analysis}) to assess the reliability and limitations of MT-Bench-X.

\subsection{MT-Bench}
\label{bench:mt_bench}
For evaluating instruction-tuned models within the conversational format, human evaluation is considered the gold standard. However, with \textit{MT-Bench}~\citep{zhengJudgingLLMasajudgeMTBench2023} a cost-effective automatic evaluation alternative, that correlates with human judgment was developed for English.
For automation \textit{MT-Bench} utilizes \glspl{LLM}-as-a-judge.
The benchmark consists of 80 high-quality, two-turn user requests across eight categories, whereas complex categories come along with reference answers.
A \gls{LLM}-as-a-judge is then prompted to assess model responses either in a pair-wise mode i.e. comparing two model responses to determine the better answer or a tie, or in a single scoring mode, where a score between 1 to 10 is to issue.
In both modes, the model is also asked to explain it's decision.
The pair-wise mode allows to check for positional bias by prompting the judge the same task twice but with reversed model response positions. 
The benchmark covers a diverse set of use cases, including \textit{Writing}, \textit{Math}, \textit{Coding}, \textit{Reasoning}, and \textit{Extraction}, among others.
To answer the research question above, we prompt GPT-4 with the judgment prompts as developed by \citet{zhengJudgingLLMasajudgeMTBench2023} but translated into the new benchmark language.
Thus, the focus of evaluation with \textit{MT-Bench} is to assess "the quality of the response provided by an AI assistant", especially in terms of "helpfulness, relevance, accuracy, depth, creativity, and level of detail of the response", as quoted from the prompt to user and machine.

\subsection{MT-Bench-X}
\label{sec:translation-mtbenchx}
Similarly to the translation of \textit{Mulima-X}, we chose DeepL as a translation engine to translate the questions, reference answers, and judge prompts of \textit{MT-Bench} from originally English to German, Spanish, Italian, and French.
Along with the original English \textit{MT-Bench}, this leads to a novel multilingual benchmark called \textit{MT-Bench-X}, which spans 400 two-turn examples in total, i.e., 80 two-turn benchmark examples per language. Publishing details are in the Appendix~\ref{appendix:licenses}.
We investigate the performance of DeepL in the Appendix~\ref{appendix:translation-problems}.
While we consider DeepL as an appropriate choice as a translation tool, there are still problematic cases that we let manually edit for correctness and amend wording for both questions and references across all languages by humans with at least a graduate
and who are fluent in the corresponding language.
For German, 31 cases were edited, many of them minor, for French 36, for Spanish 37, and for Italian 72. While French and German were correctly translated into the polite form, the Italian personal pronouns within user requests were translated into plural, which made many corrections necessary. 
Furthermore, for programming-related tasks, in some cases, variable names and control sequences are translated.
Other aspects noticeable when inspecting MT-Bench are 
the requirement of i.) translation capabilities of the \gls{LLM}-as-a-judge models, e.g. from Chinese to the translated language, and
ii.) up-to-date knowledge e.g. mentioning of GPT-4.
In addition to the user requests and references, we also translate the prompts within \textit{MT-Bench} to not mix languages systematically during evaluation with \textit{MT-Bench-X}.
Through the manual correction of the translated \textit{MT-Bench-X} dataset and the evaluation of MT-Bench-X regarding its reliability and limitations, described in the following, we offer a high-quality instruction-tuning evaluation benchmark resource to the community.

\subsection{Evaluation Approach of MT-Bench-X}
\label{sec:eval-analysis}

To assess \textit{MT-Bench-X}'s suitability, we conduct a human evaluation and validate the correlation between human and LLM-as-a-judge ratings.
We utilize the currently best model available, GPT-4, which was shown to correlate best to human evaluation for English~\citep{zhengJudgingLLMasajudgeMTBench2023}.
Furthermore, it was reported that GPT-4 is proficient in the languages we target in our study~\citep{jiaoChatGPTGoodTranslator2023,GPT4TechnicalReport2023}.
We provide a user interface inspired by \citet{zhengJudgingLLMasajudgeMTBench2023}, as can be seen in Figure \ref{fig:MT_Bench_human_eval} in the Appendix~\ref{appendix:human_eval}, to let human expert judges rate answers of models in a pair-wise response comparison. For this, we translated the prompts provided by \citet{zhengJudgingLLMasajudgeMTBench2023} into German, see Appendix~\ref{appendix:pair_wise_eval_prompts}.
Given a random question, we first set the first turn of each model response against each other and let the user choose between the options i.) Assistant A is better, ii.) Assistant B is better, iii.) Tie, iv.) both answers are not helpful or v) to skip this turn.
To reduce evaluation time, we let the second turn directly follow in the same manner.
The authors of \textit{MT-Bench}~\cite{zhengJudgingLLMasajudgeMTBench2023} minimized the risk for several biases that could occur when using a \gls{LLM} as a judge. In that spirit, we randomly select the display side for each model newly for each turn to omit positional bias.

\section{Multilingual Performance Analysis}
\label{sec:instruct-tuning-results}
To answer the question of whether polyglot models require multilingual or monolingual instructions, we conducted several experiments.
We describe the experimental setup in Section~\ref{sec:experimental-setup}, followed by the evaluation by \textit{GPT-4-as-a-judge} of single score ratings and pair-wise evaluation (Section~\ref{sec:single_score_eval}). 
Moreover, we analyze the role of dataset quality by comparing same-sized synthetic with curated datasets in Section~\ref{sec:synth_vs_curated_data} and elaborate the \textit{Superficial Alignment Hypothesis} for the multilingual setup in Section~\ref{sec:superficial_align_hypo}. 
Additionally, we conducted a qualitative analysis (Appendix ~\ref{appendix:qaulitative_analysis}).

\subsection{Experimental Setup}
\label{sec:experimental-setup}
We employ the datasets described in Section~\ref{sec:instruction-tuning} to conduct our experiments. 
This includes instruction-tuning on each monolingual dataset and all four multilingual dataset mixtures for each dataset source, i.e., \textit{Lima-X}, \textit{Bactrian-X} and \textit{Bactrian-X-small}.
As base models, we use a mid-sized multilingual 7B model that closely follows the Llama 2~\citep{touvronLlama2OpenFoundation2023} architecture, which was trained on a large number of non-English samples and the large, multilingual Mixture of Experts model \textit{Mixtral-8x7B-v0.1}. Further details are specified in the Appendix~\ref{appendix:model}.
We follow a similar hyper-parameter setting to \citet{zhouLIMALessMore2023}, i.e., we utilize a learning rate of 1e-5 with linear decrease, use a batch size of 64, utilize the Adam optimizer with $\beta_1 = 0.9$, $\beta_2 = 0.95$ and weight decay of 0.1.
Due to the high computational demand of full-weight fine-tuning of \textit{Mixtral-8x7B}, we omitted fine-tuning \textit{Mixtral-8x7B} with the large \textit{Bactrian-X} datasets. 
The number of epochs and the learning rate are determined by early stopping based on the validation loss, which resulted in a learning rate of 1e-6 for the fully parallel datasets for \textit{Mixtral-8x7B}. 

\subsection{The Impact of Language}
\label{sec:single_score_eval}
In the following, we investigate the performance of models trained on different language mixtures by both monolingual evaluation and cross-lingual evaluation.

\begin{figure*}[ht!]
    \begin{subfigure}[b]{\textwidth}
         \centering
         \includegraphics[clip, trim=1.3cm 0.2cm 1.2cm 0.1cm, width=\textwidth]{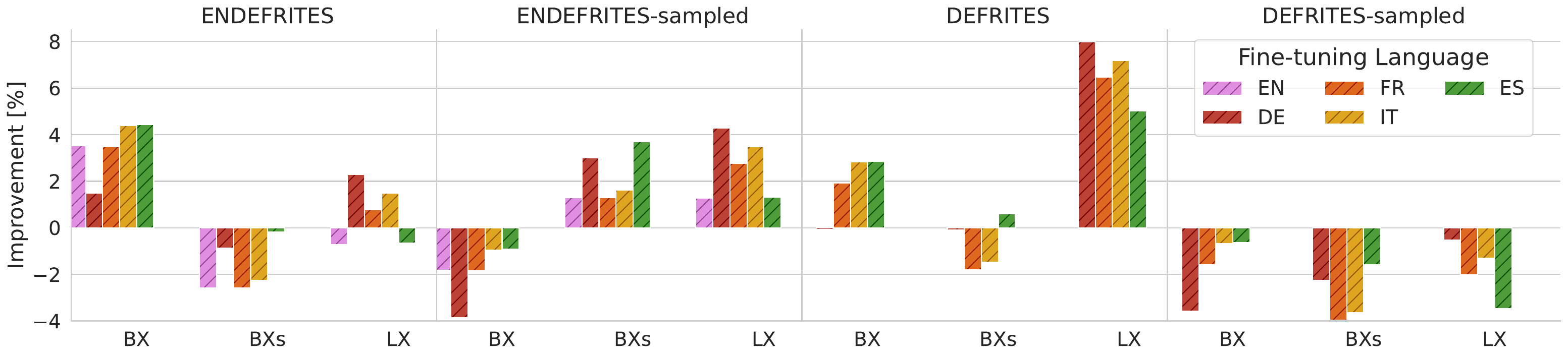}
         \caption{24EU-7B Model}
         \label{fig:avg-mt-bench-x-perc-improv-turn-1-24EU-7B}
     \end{subfigure}
     \begin{subfigure}[b]{\textwidth}
         \centering
         \includegraphics[clip, trim=1.3cm 0.2cm 1.2cm 2.9cm, width=\textwidth]{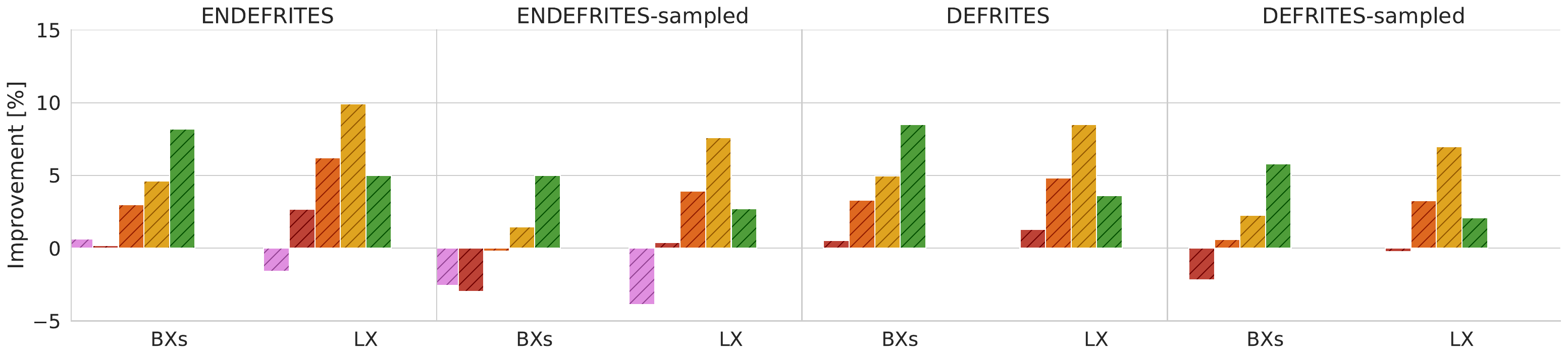}
         \caption{Mixtral-8x7B Model}
         \label{fig:avg-mt-bench-x-perc-improv-turn-1-Mixtral}
     \end{subfigure}
    \caption{Percentage improvement for turn one averaged across \textit{MT-Bench-X} languages of models fine-tuned on parallel mixed language instruction-tuning datasets over single language fine-tunings.}
    \label{fig:avg-mt-bench-x-perc-improv-turn-1}
\end{figure*}

\paragraph{Monolingual Performance}
For automatic evaluation with GPT-4, we display the summarized results within Figure~\ref{fig:mt-bench-x-catplot}, alongside fine-grained results in the Appendix~\ref{appendix:single-mt-bench-x-tables}. 
The Figure shows the average absolute results obtained on \textit{MT-Bench-X} for each instruction-tuned model.

The results for \textit{24EU-7B} in Figure~\ref{fig:mt-bench-x-catplot} highlight that the single language instruction-tunings  \texttt{Bactrian-DE}, \texttt{Bactrian-IT} and \texttt{Bactrian-ES} are within their respective evaluation language inferior compared to \texttt{Bactrian-ENDEFRITES}.
In contrast, \texttt{Bactrian-EN} achieves strong performance, as Figure~\ref{fig:mt-bench-x-catplot} shows. We hypothesize this is due to the large fraction of English data within the pre-training corpus.
Also, \texttt{Bactrian-FR} shows to outperform \texttt{Bactrian-ENDEFRITES} within \textit{MT-Bench-FR}.
The fully parallel \textit{DEFRITES}-based and \textit{ENDEFRITES}-based language mixture fine-tunings show no consistent superiority. 
\textit{Bactrian-X-small} shows a coarsely similar, albeit overall lower performance compared to  \textit{Bactrian-X}.

\textit{Mixtral-7x8B}-based models showcase an overall higher absolute score within each language, compared to \textit{24EU-7B}-based models, supposedly due to its larger size.
For the \textit{Mixtral-7x8B}-based results, a consistent effect of monolingual performance increase or at least comparative performance for instruction-tuning with cross-lingual datasets compared to monolingual datasets can be observed for \textit{Bactrian-X-small} as well as \textit{Lima-X}-based models. However, as with the \textit{24EU-7B} model results, either \textit{DEFRITES} or \textit{ENDEFRITES} language mixture based models show this effect inconsistently. 
There is no clear picture when evaluating model performance in a single language only.

\paragraph{Cross-lingual Performance}
We compare the average performance across languages between parallel dataset mixtures and monolingual datasets in Figure~\ref{fig:avg-mt-bench-x-perc-improv-turn-1}. 
As can be inferred from Figure~\ref{fig:avg-mt-bench-x-perc-improv-turn-1-24EU-7B}, for \textit{24EU-7B}-based models the language mix \texttt{ENDEFRITES-sampled}, multilingual instruction-tuning improvements for \textit{Lima-X} and the same sized \textit{Bactrian-X-small} are notable, but the opposite for \textit{Bactrian-X}. However, the same cannot be observed for \textit{DEFRITES-sampled}, as here, all models trained with parallel data mixtures are inferior compared to their monolingual counterparts.
The inconsistency within these results might come from the number of samples per language, which is five times as small within \textit{ENDEFRITES-sampled} and \textit{DEFRITES-sampled} compared to the full monolingual dataset. Here, \textit{Lima-X} and \textit{Bactrian-X-small} only contain 1030 samples in total, i.e. 206 samples per language within \texttt{ENDEFRITES-sampled}.

For \textit{Mixtral-8x7B} models in Figure~\ref{fig:avg-mt-bench-x-perc-improv-turn-1-Mixtral}, we note consistent improvement of parallel instruction-tuning, i.e., for the sampled as well as full-sized datasets, with only a few, but consistent and less pronounced exceptions for the same language and dataset combination.
In the same manner, the \textit{24EU-7B} models trained based on the full-sized, parallel language mix \texttt{ENDEFRITES} and \textit{DEFRITES} show performance improvements across languages for \textit{Bactrian-X} and \textit{Lima-X}, when compared to their monolingual counterparts.

Furthermore, the performance improvement when including the predominant pre-training language during parallel instruction-tuning (\texttt{ENDEFRITES} vs \texttt{DEFRITES}) highlights the importance of its inclusion during instruction-tuning.
However, the down-sampled variants show decreased performances for both parallel language mixtures. We assume this is due to the synthetic nature of \textit{Bactrian-X}, as it may show different levels of hallucination depending on the language skill of the generative model. Therefore, the sampled setting could have a lower signal-to-noise ratio than the \texttt{ENDEFRITES} setting. In contrast, LIMA-sampled, a carefully curated dataset, does not express a performance decrease, providing evidence for the above-mentioned hypothesis.
We assume the increased world knowledge associated with a larger model size allows \textit{Mixtral-8x7B} to compensate for the assumed low signal-to-noise ratio in \textit{Bactrian-X-small}.

We observe a maximum performance gain of 9.9\% (0.99 points on the MT-Bench scale ranging from 1 to 10) by \texttt{Bactrian-ENDEFRITES} compared to \texttt{Bactrian-IT}.

Despite the emergence of multilingual capabilities during pre-training, we found fine-tuning on parallel data improves multilingual instruction-following performance stronger across languages compared to non-parallel multilingual data as well as monolingual data for mid-sized and large \glspl{LLM}.

\subsection{Synthetic vs Curated Training Datasets}
\label{sec:synth_vs_curated_data}
To factor out the impact of training dataset size, we down-sampled the \textit{Bactrian-X} datasets to the same size as the human-curated \textit{LIMA-X} datasets, resulting in the dataset \textit{Bactrian-X-small}. As shown in Figure~\ref{fig:mt-bench-x-catplot} in the Appendix, the Bactrian-X-based synthetic datasets outperform \textit{LIMA-X} for both dataset magnitudes.
When comparing \textit{LIMA-X} with the cross-lingual performance of models tuned on \textit{Bactrian-X-small} or \textit{Bactrian-X} in Figure~\ref{fig:avg-mt-bench-x-perc-improv-turn-1}, the same overall performance trend i.e. improved cross-lingual performance for parallel Bactrian-X-based instruction-tunings, is visible.
These results underline the beneficial effect of employing synthetic, semantically parallel datasets for multilingual instruction-tuning.

\subsection{Superficial Alignment Hypothesis}
\label{sec:superficial_align_hypo}
We additionally investigated the superficial alignment hypothesis, which states that only a few examples per task are required to teach a model to follow instructions.
Figure~\ref{fig:mt-bench-x-catplot} in the Appendix highlights that \textit{24EU-7B} models instruction-tuned on \textit{Bactrian-X} as well as \textit{Bactrian-X-small} generally outperform models instruction-tuned on \textit{Lima-X} datasets. 
Our results show that the \textit{Superficial Alignment Hypothesis}~\citep{kirstain-etal-2022-examples, zhouLIMALessMore2023} does not generally hold for mid-sized \glspl{LLM}.
However, with \textit{Mixtral-8x7B}, we see high performances for synthetic as well as human-curated data, indicating that the effectiveness of the \textit{Superficial Alignment Hypothesis} increases with larger model size or respectively with greater pre-trained model capabilities.

\section{Human Evaluation}
\label{sec:pair_wise_human_eval}
Human evaluation is the gold standard for evaluating the output of generative models, as responses can be complex and highly diverse.
Following the human evaluation as described in Section~\ref{sec:eval-analysis} on answers of the \textit{24EU-7B} model tuned on \textit{Bactrian-X}, we:
i.) inspect the role of positional bias in \textit{MT-Bench-X} (Section~\ref{sec:pos_bias_main}),
ii.) compare the voting distributions between humans and GPT-4 in Section~\ref{sec:vote_distr},
and iii.) analyze the correlation of agreement levels between humans and GPT-4 in Section~\ref{sec:corr_analysis}.

\subsection{Positional Bias}
\label{sec:pos_bias_main}
We inspect potential limitations of utilizing \textit{GPT-4-as-a-judge} exemplary for German text in Table~\ref{tab:pos-bias}.
We observe a high level of positional bias for the categories \textit{Stem}, \textit{Humanities} and \textit{Writing}.
For the following correlation analysis, we mitigate the effect of positional bias, in which case no result is provided for a given benchmark example, by substituting missing values through the results of the following run, where possible.
Albeit judgment generation in \textit{MT-Bench-X} is conducted by greedy search and the evaluation runs were executed immediately after another, we mitigated the positional bias by up to 16.25\%.

\begin{table}[ht!]
\centering
   \begin{tabular}{lrlr}
   \toprule
    Category & PB [\%] & Category & PB [\%]\\
     \midrule
    Stem & 30.00 & Roleplay & 15.00 \\
    Humanities & 27.50 & Coding & 12.50 \\
    Writing & 20.00 & Reasoning & 7.50 \\
    Extraction & 16.25 & Math & 6.25 \\
    \bottomrule
\end{tabular}
    \caption{Average percentage of positional bias (PB).}
    \label{tab:pos-bias}
\end{table}

\begin{figure*}[t!]
     \centering
    \begin{subfigure}[b]{0.495\textwidth}
         \centering
         \includegraphics[clip, trim=0cm 0cm 0cm 0.6cm, width=\textwidth]{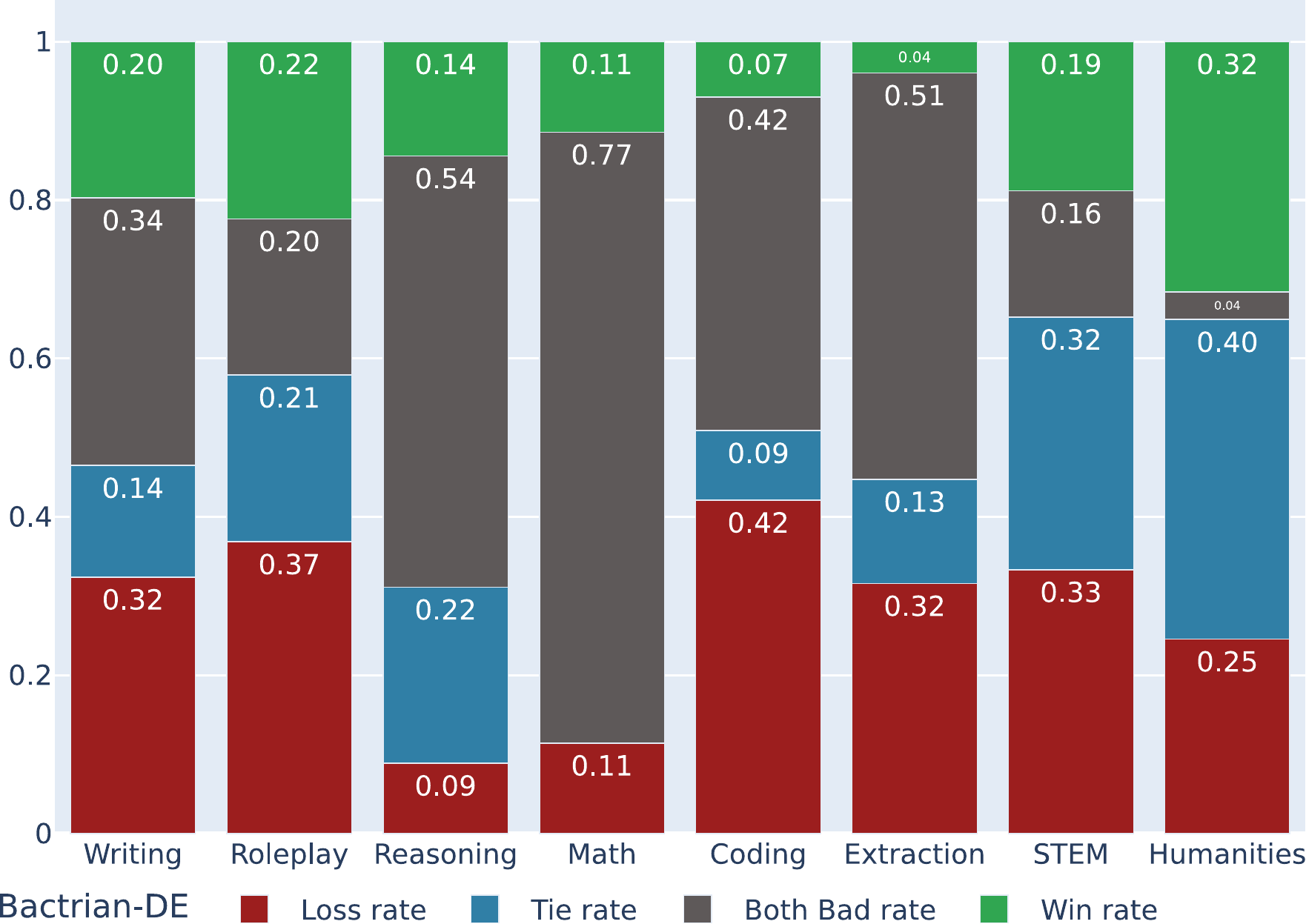}
        \caption{\texttt{Bactrian-DE} vs.  \texttt{Bactrian-ENDEFRITES} voted by humans.
        }
        \label{fig:pair-wise-gpt-4-mt-bench-DE-both-bad-human-eval}
    \end{subfigure}
    \begin{subfigure}[b]{0.495\textwidth}
         \centering
         \includegraphics[clip, trim=0cm 0cm 0cm 0.6cm, width=\textwidth]{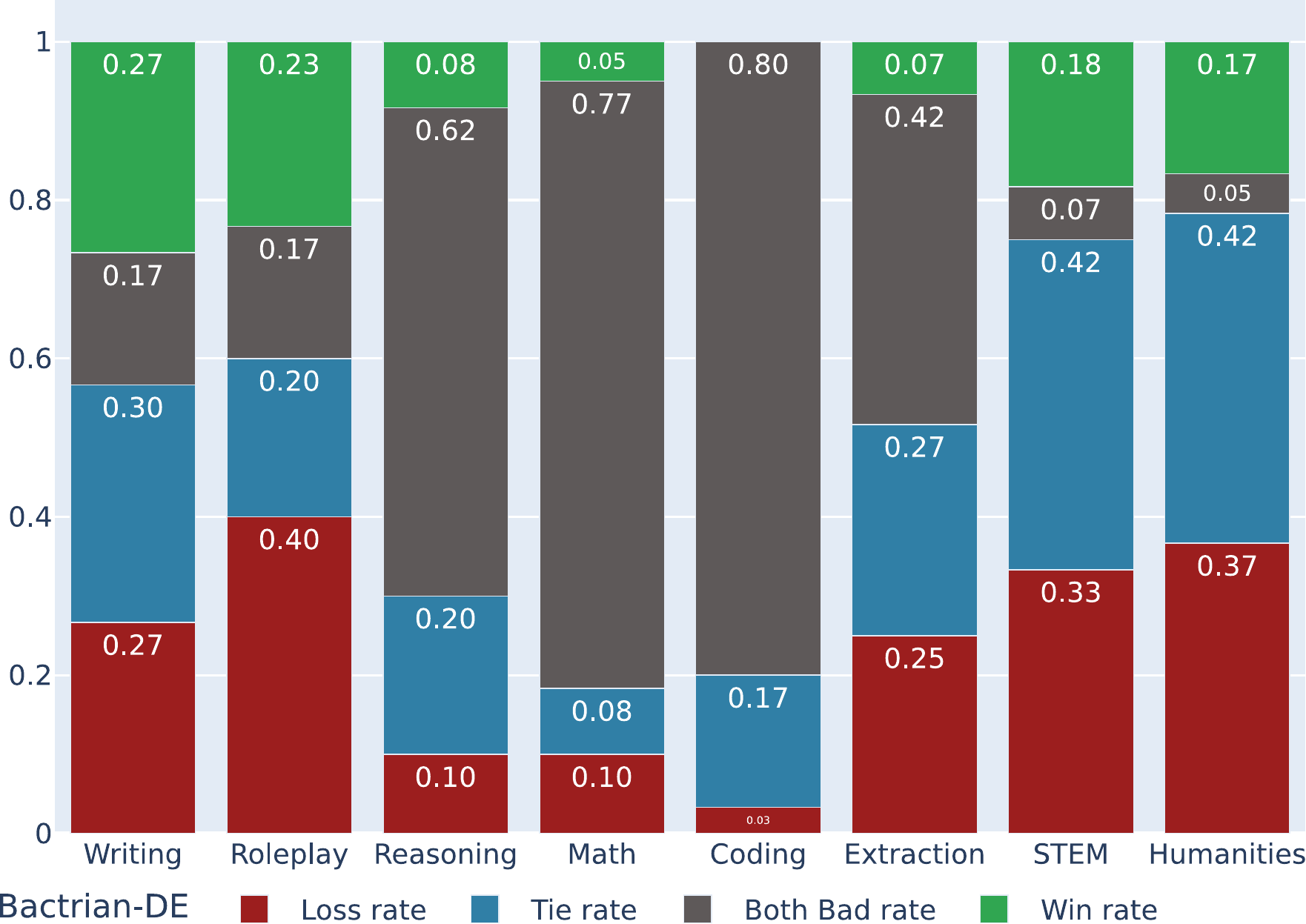}
        \caption{\texttt{Bactrian-DE} vs.  \texttt{Bactrian-ENDEFRITES} 
        voted by GPT-4.
        }
        \label{fig:pair-wise-gpt-4-mt-bench-DE-both-bad}
    \end{subfigure}
    \caption{Pair-wise MT-Bench-DE quality assessment by humans and GPT-4, including voting option "both bad".}
    \label{fig:gpt_4_mt_bench_DE_pair_human_and_machine_vo}
\end{figure*}
\subsection{Voting Distribution}
\label{sec:vote_distr}
We averaged the results of three evaluation runs for the models \texttt{Bactrian-DE} and \texttt{Bactrian-ENDEFRITES} on \textit{MT-Bench-DE} and display the results in Figure~\ref{fig:gpt_4_mt_bench_DE_pair_human_and_machine_vo}.
As can be seen in Figure~\ref{fig:pair-wise-gpt-4-mt-bench-DE-both-bad}, \texttt{Bactrian-DE} loses more frequently against \texttt{Bactrian-ENDEFRITES}, except for the \textit{Reasoning} category.     
Especially for the categories \textit{Math}, \textit{Reasoning}, and \textit{Coding}, the model performance is insufficient, and thus, a performance comparison is infeasible. We hypothesize this shows a gap in capabilities learned during pre-training.
Within Figure~\ref{fig:pair-wise-gpt-4-mt-bench-DE-both-bad-human-eval} it is evident that human evaluators tend to vote less often for "Tie" and "Both Bad".
For the categories \textit{Math}, \textit{Reasoning} and \textit{Coding}, the underperformance shown by automatic per-category single evaluation scores in Figure~\ref{fig:radar-gpt-4-mt-bench-X} in the Appendix is confirmed by the human evaluation in Figure~\ref{fig:pair-wise-gpt-4-mt-bench-DE-both-bad-human-eval}.

\begin{table}[h!]
    \centering
    \begin{threeparttable}
        \begin{tabular}{lrrrr}
            \toprule
                                                                                                                                                                    
                                                Turn & \multicolumn{2}{c}{1} & \multicolumn{2}{c}{2} \\
            Category & GPT-4 & Hum. & GPT-4 & Hum. \\
            \midrule
            All & $39_{59}$ & 56 & $40_{62}$ & 52 \\
            \textit{RCEMR}\tnote{a} & $50_{60}$ & 57 & $46_{71}$ & 58 \\
            Coding & $47_{80}$ & 61 & $11_{36}$ & 33 \\
            Extraction & $51_{48}$ & 48 & $29_{81}$ & 62 \\
            Humanities & $0_{88}$ & 77 & $44_{46}$ & 66 \\
            Math & $62_{92}$ & 85 & $70_{88}$ & 77 \\
            Reasoning & $43_{54}$ & 60 & $70_{83}$ & 66 \\
            Roleplay & $46_{45}$ & 33 & $33_{70}$ & 41 \\
            Stem & $20_{60}$ & 54 & $26_{42}$ & 26 \\
            Writing & $16_{49}$ & 46 & $18_{51}$ & 29 \\
            \bottomrule
            \end{tabular}
        \smallskip
        \footnotesize
        \begin{tablenotes}
        \item[a] \textit{RCEMR}: Roleplay, Coding, Extraction, Math and Reasoning
        \end{tablenotes}
    \end{threeparttable}
    \caption{Percentage of agreement between Human-GPT-4 and Human-Human with subscripted upper limit of Human-Majority-Human agreement.}
    \label{tab:corr-analysis-results}
\end{table}
\subsection{Correlation Analysis}
\label{sec:corr_analysis}
The correlation between agreement levels of humans and GPT-4 is shown in Table~\ref{tab:corr-analysis-results}. 
As can be seen, there is an agreement of 39\% between humans and GPT-4 for turn one and 56\% agreement between humans.
To incorporate the agreement by chance, we calculate the Fleiss' kappa scores~\cite{fleiss1973equivalence} among humans, resulting in the coefficient of 0.37 (p-value: $\num{3e-8}$) on 66 samples for the first turn and 0.36 (p-value: $\num{6e-8}$) on 62 samples for the second turn. This indicates fair agreement between annotators.
We assume the low scores are effects of the difficulty of the evaluation task itself, as even human judges agree only to 56\% to each other.
Due to the disagreement among human judges only the upper bound of 59\% agreement can be reached for human-GPT-4 agreement.
Interestingly, \textit{Humanities}, \textit{Writing} and \textit{Stem} contribute significantly to the disagreement level between humans and GPT-4. We attribute this to the positional bias, which was especially observable within categories that involve creativity and thus are more subjective to assess.
Of the 242 model responses, human judges prefer to 62.8\% the model responses of \texttt{Bactrian-ENDEFRITES}, which further supports our findings in Sec. \ref{sec:single_score_eval}.

\section{Conclusion \& Future Work}

We see this work as a fundamental step towards supportive multilingual assistants.
Across a selection of major Indo-European languages, we thoroughly analyzed instruction-tuned models using parallel instruction-tuning benchmarks. Hereby, the creation of multilingual benchmarks and datasets are crucial to investigate multilingual instruction-tuning capabilities.
Our findings highlight the benefits of instruction-tuning on parallel datasets, showcasing improvements of up to 9.9\% in cross-lingual instruction-following capabilities compared to monolingual corpora. Additionally, our results challenge the generalizability of the \textit{Superficial Alignment Hypothesis}, showing that extensive instruction-tuning datasets are necessary for mid-sized multilingual models, but not necessarily for larger models.
We extended the findings of \citet{zhengJudgingLLMasajudgeMTBench2023} for German and analyzed the similarities and differences between human evaluation and evaluation with GPT-4-as-a-judge.
We identify disparities between human evaluations and those generated by GPT-4 in multilingual chat scenarios.

By illuminating these challenges, we emphasize the need for future research to address them.
Additionally, we recognize the need to explore the impact of multilingual multi-turn dataset variants, which we leave as an avenue for future exploration.
By addressing these challenges head-on, we can improve the performance of generative assistants in real-world communication contexts, advancing the field of natural language processing for practical applications.

\section{Limitations}
While our study offers valuable insights into instruction-tuning for multilingual \glspl{LLM}, it is essential to acknowledge the following limitations.

Firstly, our research does not aim to push the boundaries of state-of-the-art performance. Instead, we focus on exploring the effectiveness of different instruction-tuning settings in guiding pre-trained multilingual \glspl{LLM} to follow instructions.

Secondly, due to involved costs when evaluating with \textit{GPT-4-as-a-judge}, we conducted single-score evaluations for each model variant across various languages in the \textit{MT-Bench-X} dataset only once. While this approach provided initial insights, it limited our ability to calculate comprehensive statistical measures like mean and standard deviation.

Moreover, our research scope is confined to languages within the Germanic and Italo-Western language families due to resource constraints. Consequently, the generalizability of our findings to languages from more distant language families remains to be determined.
Despite these limitations, our study lays the groundwork for exploring whether multilingual instruction-tuning benefits languages beyond those examined in this research, opening avenues for further investigation of multilingual \gls{LLM} fine-tuning methodologies.

\section{Ethical And Broader Impact}
Instruction-following \glspl{LLM} offers an efficient way of solving natural language problems by simply instructing the model to perform the tasks.
With our work, we highlight the importance of investigating the multilingual aspect throughout the creation process of helpful \glspl{LLM}, as this becomes an important feature for democratizing this technology.
While this allows users to become proficient in various areas, pre-trained and instruction-tuned models are not restricted out-of-the-box to a certain set of content and do not follow a specific set of values.
Thus an important next step is to investigate the generalizability of the alignment to human curated values embedded within moderated datasets across multiple languages.

\section*{Acknowledgments}
We would like to thank Dr. Joachim Köhler, Ines Wendler, Joe Jamison, and Valentina Ciardini (Fraunhofer IIS) for their invaluable support for insightful discussions and participation for the quality assessment of created resources. 
We would like to extend our gratitude to the Fraunhofer IAIS team for their valuable contributions to this project, particularly their involvement in human evaluation.
This work was funded by the Federal Ministry of Education and Research of Germany and the state of North-Rhine Westphalia as part of the Lamarr-Institute for Machine Learning and Artificial Intelligence, LAMARR22B as well as by the German Federal Ministry for Economic Affairs and Climate Action (BMWK) through the project OpenGPT-X (project no. 68GX21007D) and by the European Union’s Horizon 2020 research and innovation program under grant agreement No 101135671 (TrustLLM).
The authors gratefully acknowledge the Gauss Centre for Supercomputing e.V. (www.gauss-centre.eu) for funding this project by providing computing time on the GCS Supercomputer JUWELS at Jülich Supercomputing Centre (JSC) as well as the Center for Information Services and High Performance Computing [Zentrum für Informationsdienste und Hochleistungsrechnen (ZIH)] at TU Dresden for providing its facilities for automatic evaluation computations.

\bibliography{main}

\begin{thebibliography}{35}
\expandafter\ifx\csname natexlab\endcsname\relax\def\natexlab#1{#1}\fi

\bibitem[{Ali et~al.(2024)Ali, Fromm, Thellmann, Ebert, Weber, Rutmann, Jain, Lübbering, Steinigen, Leveling, Klug, Buschhoff, Jurkschat, Abdelwahab, Stein, Sylla, Denisov, Brandizzi, Saleem, Anirban, John, Suarez, Ostendorff, Jude, Manjunath, Weinbach, Penke, Asaadi, Barth, Sifa, Küch, Jäkel, Rehm, Kesselheim, Köhler, and Flores-Herr}]{ali2024progressreporteuropeanllms}
Mehdi Ali, Michael Fromm, Klaudia Thellmann, Jan Ebert, Alexander~Arno Weber, Richard Rutmann, Charvi Jain, Max Lübbering, Daniel Steinigen, Johannes Leveling, Katrin Klug, Jasper~Schulze Buschhoff, Lena Jurkschat, Hammam Abdelwahab, Benny~Jörg Stein, Karl-Heinz Sylla, Pavel Denisov, Nicolo Brandizzi, Qasid Saleem, Bhowmick Anirban, Chelsea John, Pedro~Ortiz Suarez, Malte Ostendorff, Alex Jude, Lalith Manjunath, Samuel Weinbach, Carolin Penke, Shima Asaadi, Fabio Barth, Rafet Sifa, Fabian Küch, René Jäkel, Georg Rehm, Stefan Kesselheim, Joachim Köhler, and Nicolas Flores-Herr. 2024.
\newblock \href {http://arxiv.org/abs/2410.03730} {Progress report: Towards european llms}.

\bibitem[{Ali et~al.(2023)Ali, Fromm, Thellmann, Rutmann, L{\"{u}}bbering, Leveling, Klug, Ebert, Doll, Buschhoff, Jain, Weber, Jurkschat, Abdelwahab, John, Suarez, Ostendorff, Weinbach, Sifa, Kesselheim, and Flores{-}Herr}]{aliTokenizerChoiceLLM2023}
Mehdi Ali, Michael Fromm, Klaudia Thellmann, Richard Rutmann, Max L{\"{u}}bbering, Johannes Leveling, Katrin Klug, Jan Ebert, Niclas Doll, Jasper~Schulze Buschhoff, Charvi Jain, Alexander~Arno Weber, Lena Jurkschat, Hammam Abdelwahab, Chelsea John, Pedro~Ortiz Suarez, Malte Ostendorff, Samuel Weinbach, Rafet Sifa, Stefan Kesselheim, and Nicolas Flores{-}Herr. 2023.
\newblock \href {https://doi.org/10.48550/ARXIV.2310.08754} {Tokenizer choice for {LLM} training: Negligible or crucial?}
\newblock \emph{CoRR}, abs/2310.08754.

\bibitem[{Bai et~al.(2022)Bai, Jones, Ndousse, Askell, Chen, DasSarma, Drain, Fort, Ganguli, Henighan, Joseph, Kadavath, Kernion, Conerly, Showk, Elhage, Hatfield{-}Dodds, Hernandez, Hume, Johnston, Kravec, Lovitt, Nanda, Olsson, Amodei, Brown, Clark, McCandlish, Olah, Mann, and Kaplan}]{baiTrainingHelpfulHarmless2022}
Yuntao Bai, Andy Jones, Kamal Ndousse, Amanda Askell, Anna Chen, Nova DasSarma, Dawn Drain, Stanislav Fort, Deep Ganguli, Tom Henighan, Nicholas Joseph, Saurav Kadavath, Jackson Kernion, Tom Conerly, Sheer~El Showk, Nelson Elhage, Zac Hatfield{-}Dodds, Danny Hernandez, Tristan Hume, Scott Johnston, Shauna Kravec, Liane Lovitt, Neel Nanda, Catherine Olsson, Dario Amodei, Tom~B. Brown, Jack Clark, Sam McCandlish, Chris Olah, Benjamin Mann, and Jared Kaplan. 2022.
\newblock \href {https://doi.org/10.48550/ARXIV.2204.05862} {Training a helpful and harmless assistant with reinforcement learning from human feedback}.
\newblock \emph{CoRR}, abs/2204.05862.

\bibitem[{Ding et~al.(2023)Ding, Chen, Xu, Qin, Hu, Liu, Sun, and Zhou}]{DBLP:conf/emnlp/DingCXQHL0Z23ultrachat}
Ning Ding, Yulin Chen, Bokai Xu, Yujia Qin, Shengding Hu, Zhiyuan Liu, Maosong Sun, and Bowen Zhou. 2023.
\newblock \href {https://doi.org/10.18653/V1/2023.EMNLP-MAIN.183} {Enhancing chat language models by scaling high-quality instructional conversations}.
\newblock In \emph{Proceedings of the 2023 Conference on Empirical Methods in Natural Language Processing, {EMNLP} 2023, Singapore, December 6-10, 2023}, pages 3029--3051. Association for Computational Linguistics.

\bibitem[{Ferron et~al.(2023)Ferron, Shore, Mitra, and Agrawal}]{ferronMEEPThisEngaging2023}
Amila Ferron, Amber Shore, Ekata Mitra, and Ameeta Agrawal. 2023.
\newblock \href {https://aclanthology.org/2023.findings-emnlp.137} {{MEEP:} is this engaging? prompting large language models for dialogue evaluation in multilingual settings}.
\newblock In \emph{Findings of the Association for Computational Linguistics: {EMNLP} 2023, Singapore, December 6-10, 2023}, pages 2078--2100. Association for Computational Linguistics.

\bibitem[{Fleiss and Cohen(1973)}]{fleiss1973equivalence}
Joseph~L Fleiss and Jacob Cohen. 1973.
\newblock The equivalence of weighted kappa and the intraclass correlation coefficient as measures of reliability.
\newblock \emph{Educational and psychological measurement}, 33(3):613--619.

\bibitem[{Hidalgo-Ternero(2020)}]{hidalgo2020google}
Carlos~Manuel Hidalgo-Ternero. 2020.
\newblock Google translate vs. deepl.
\newblock \emph{MonTI. Monograf{\'\i}as de Traducci{\'o}n e Interpretaci{\'o}n}, pages 154--177.

\bibitem[{Holmstr{\"{o}}m and Doostmohammadi(2023)}]{holmstromMakingInstructionFinetuningSwedish2023}
Oskar Holmstr{\"{o}}m and Ehsan Doostmohammadi. 2023.
\newblock \href {https://aclanthology.org/2023.nodalida-1.62} {Making instruction finetuning accessible to non-english languages: {A} case study on swedish models}.
\newblock In \emph{Proceedings of the 24th Nordic Conference on Computational Linguistics, NoDaLiDa 2023, T{\'{o}}rshavn, Faroe Islands, May 22-24, 2023}, pages 634--642. University of Tartu Library.

\bibitem[{Iyer et~al.(2022)Iyer, Lin, Pasunuru, Mihaylov, Simig, Yu, Shuster, Wang, Liu, Koura, Li, O'Horo, Pereyra, Wang, Dewan, Celikyilmaz, Zettlemoyer, and Stoyanov}]{iyerOPTIMLScalingLanguage2022}
Srinivasan Iyer, Xi~Victoria Lin, Ramakanth Pasunuru, Todor Mihaylov, Daniel Simig, Ping Yu, Kurt Shuster, Tianlu Wang, Qing Liu, Punit~Singh Koura, Xian Li, Brian O'Horo, Gabriel Pereyra, Jeff Wang, Christopher Dewan, Asli Celikyilmaz, Luke Zettlemoyer, and Ves Stoyanov. 2022.
\newblock \href {https://doi.org/10.48550/ARXIV.2212.12017} {{OPT-IML:} scaling language model instruction meta learning through the lens of generalization}.
\newblock \emph{CoRR}, abs/2212.12017.

\bibitem[{Jiang et~al.(2024)Jiang, Sablayrolles, Roux, Mensch, Savary, Bamford, Chaplot, de~Las~Casas, Hanna, Bressand, Lengyel, Bour, Lample, Lavaud, Saulnier, Lachaux, Stock, Subramanian, Yang, Antoniak, Scao, Gervet, Lavril, Wang, Lacroix, and Sayed}]{jiangMixtralExperts2024}
Albert~Q. Jiang, Alexandre Sablayrolles, Antoine Roux, Arthur Mensch, Blanche Savary, Chris Bamford, Devendra~Singh Chaplot, Diego de~Las~Casas, Emma~Bou Hanna, Florian Bressand, Gianna Lengyel, Guillaume Bour, Guillaume Lample, L{\'{e}}lio~Renard Lavaud, Lucile Saulnier, Marie{-}Anne Lachaux, Pierre Stock, Sandeep Subramanian, Sophia Yang, Szymon Antoniak, Teven~Le Scao, Th{\'{e}}ophile Gervet, Thibaut Lavril, Thomas Wang, Timoth{\'{e}}e Lacroix, and William~El Sayed. 2024.
\newblock \href {https://doi.org/10.48550/ARXIV.2401.04088} {Mixtral of experts}.
\newblock \emph{CoRR}, abs/2401.04088.

\bibitem[{Jiao et~al.()Jiao, Wang, Huang, Wang, Shi, and Tu}]{jiaoChatGPTGoodTranslator2023}
Wenxiang Jiao, Wenxuan Wang, Jen-tse Huang, Xing Wang, Shuming Shi, and Zhaopeng Tu.
\newblock \href {https://doi.org/10.48550/arXiv.2301.08745} {Is {{ChatGPT A Good Translator}}? {{Yes With GPT-4 As The Engine}}}.

\bibitem[{Kirstain et~al.(2022)Kirstain, Lewis, Riedel, and Levy}]{kirstain-etal-2022-examples}
Yuval Kirstain, Patrick Lewis, Sebastian Riedel, and Omer Levy. 2022.
\newblock \href {https://doi.org/10.18653/v1/2022.findings-emnlp.72} {A few more examples may be worth billions of parameters}.
\newblock In \emph{Findings of the Association for Computational Linguistics: EMNLP 2022}, pages 1017--1029, Abu Dhabi, United Arab Emirates. Association for Computational Linguistics.

\bibitem[{K{\"{o}}pf et~al.(2023)K{\"{o}}pf, Kilcher, von R{\"{u}}tte, Anagnostidis, Tam, Stevens, Barhoum, Duc, Stanley, Nagyfi, ES, Suri, Glushkov, Dantuluri, Maguire, Schuhmann, Nguyen, and Mattick}]{kopfOpenAssistantConversationsDemocratizing2023}
Andreas K{\"{o}}pf, Yannic Kilcher, Dimitri von R{\"{u}}tte, Sotiris Anagnostidis, Zhi{-}Rui Tam, Keith Stevens, Abdullah Barhoum, Nguyen~Minh Duc, Oliver Stanley, Rich{\'{a}}rd Nagyfi, Shahul ES, Sameer Suri, David Glushkov, Arnav Dantuluri, Andrew Maguire, Christoph Schuhmann, Huu Nguyen, and Alexander Mattick. 2023.
\newblock \href {https://doi.org/10.48550/ARXIV.2304.07327} {Openassistant conversations - democratizing large language model alignment}.
\newblock \emph{CoRR}, abs/2304.07327.

\bibitem[{Kudo(2018)}]{kudoUnigram2018}
Taku Kudo. 2018.
\newblock \href {https://doi.org/10.18653/V1/P18-1007} {Subword regularization: Improving neural network translation models with multiple subword candidates}.
\newblock In \emph{Proceedings of the 56th Annual Meeting of the Association for Computational Linguistics, {ACL} 2018, Melbourne, Australia, July 15-20, 2018, Volume 1: Long Papers}, pages 66--75. Association for Computational Linguistics.

\bibitem[{Lai et~al.(2023)Lai, Nguyen, Ngo, Nguyen, Dernoncourt, Rossi, and Nguyen}]{laiOkapiInstructiontunedLarge2023}
Viet~Dac Lai, Chien~Van Nguyen, Nghia~Trung Ngo, Thuat Nguyen, Franck Dernoncourt, Ryan~A. Rossi, and Thien~Huu Nguyen. 2023.
\newblock \href {https://aclanthology.org/2023.emnlp-demo.28} {Okapi: Instruction-tuned large language models in multiple languages with reinforcement learning from human feedback}.
\newblock In \emph{Proceedings of the 2023 Conference on Empirical Methods in Natural Language Processing, {EMNLP} 2023 - System Demonstrations, Singapore, December 6-10, 2023}, pages 318--327. Association for Computational Linguistics.

\bibitem[{Li et~al.(2023)Li, Koto, Wu, Aji, and Baldwin}]{liBactrianXMultilingualReplicable2023}
Haonan Li, Fajri Koto, Minghao Wu, Alham~Fikri Aji, and Timothy Baldwin. 2023.
\newblock \href {https://doi.org/10.48550/ARXIV.2305.15011} {Bactrian-x : {A} multilingual replicable instruction-following model with low-rank adaptation}.
\newblock \emph{CoRR}, abs/2305.15011.

\bibitem[{Liu et~al.(2023)Liu, Iter, Xu, Wang, Xu, and Zhu}]{liuGEvalNLGEvaluation2023}
Yang Liu, Dan Iter, Yichong Xu, Shuohang Wang, Ruochen Xu, and Chenguang Zhu. 2023.
\newblock \href {https://aclanthology.org/2023.emnlp-main.153} {G-eval: {NLG} evaluation using gpt-4 with better human alignment}.
\newblock In \emph{Proceedings of the 2023 Conference on Empirical Methods in Natural Language Processing, {EMNLP} 2023, Singapore, December 6-10, 2023}, pages 2511--2522. Association for Computational Linguistics.

\bibitem[{Longpre et~al.(2023)Longpre, Hou, Vu, Webson, Chung, Tay, Zhou, Le, Zoph, Wei, and Roberts}]{longpreFlanCollectionDesigning2023}
Shayne Longpre, Le~Hou, Tu~Vu, Albert Webson, Hyung~Won Chung, Yi~Tay, Denny Zhou, Quoc~V. Le, Barret Zoph, Jason Wei, and Adam Roberts. 2023.
\newblock \href {https://proceedings.mlr.press/v202/longpre23a.html} {The flan collection: Designing data and methods for effective instruction tuning}.
\newblock In \emph{International Conference on Machine Learning, {ICML} 2023, 23-29 July 2023, Honolulu, Hawaii, {USA}}, volume 202 of \emph{Proceedings of Machine Learning Research}, pages 22631--22648. {PMLR}.

\bibitem[{Mishra et~al.(2022)Mishra, Khashabi, Baral, and Hajishirzi}]{DBLP:conf/acl/MishraKBH22naturalinstruct}
Swaroop Mishra, Daniel Khashabi, Chitta Baral, and Hannaneh Hajishirzi. 2022.
\newblock \href {https://doi.org/10.18653/V1/2022.ACL-LONG.244} {Cross-task generalization via natural language crowdsourcing instructions}.
\newblock In \emph{Proceedings of the 60th Annual Meeting of the Association for Computational Linguistics (Volume 1: Long Papers), {ACL} 2022, Dublin, Ireland, May 22-27, 2022}, pages 3470--3487. Association for Computational Linguistics.

\bibitem[{Muennighoff et~al.(2023)Muennighoff, Wang, Sutawika, Roberts, Biderman, Scao, Bari, Shen, Yong, Schoelkopf, Tang, Radev, Aji, Almubarak, Albanie, Alyafeai, Webson, Raff, and Raffel}]{muennighoffCrosslingualGeneralizationMultitask2023}
Niklas Muennighoff, Thomas Wang, Lintang Sutawika, Adam Roberts, Stella Biderman, Teven~Le Scao, M.~Saiful Bari, Sheng Shen, Zheng~Xin Yong, Hailey Schoelkopf, Xiangru Tang, Dragomir Radev, Alham~Fikri Aji, Khalid Almubarak, Samuel Albanie, Zaid Alyafeai, Albert Webson, Edward Raff, and Colin Raffel. 2023.
\newblock \href {https://doi.org/10.18653/V1/2023.ACL-LONG.891} {Crosslingual generalization through multitask finetuning}.
\newblock In \emph{Proceedings of the 61st Annual Meeting of the Association for Computational Linguistics (Volume 1: Long Papers), {ACL} 2023, Toronto, Canada, July 9-14, 2023}, pages 15991--16111. Association for Computational Linguistics.

\bibitem[{OpenAI(2023)}]{GPT4TechnicalReport2023}
OpenAI. 2023.
\newblock \href {https://doi.org/10.48550/ARXIV.2303.08774} {{GPT-4} technical report}.
\newblock \emph{CoRR}, abs/2303.08774.

\bibitem[{Petrov et~al.(2023)Petrov, Malfa, Torr, and Bibi}]{petrovLanguageModelTokenizers2023}
Aleksandar Petrov, Emanuele~La Malfa, Philip H.~S. Torr, and Adel Bibi. 2023.
\newblock Language model tokenizers introduce unfairness between languages.
\newblock \emph{CoRR}, abs/2305.15425.

\bibitem[{Ponti et~al.(2020)Ponti, Glava{\v{s}}, Majewska, Liu, Vuli{\'c}, and Korhonen}]{ponti-etal-2020-xcopa}
Edoardo~Maria Ponti, Goran Glava{\v{s}}, Olga Majewska, Qianchu Liu, Ivan Vuli{\'c}, and Anna Korhonen. 2020.
\newblock \href {https://doi.org/10.18653/v1/2020.emnlp-main.185} {{XCOPA}: A multilingual dataset for causal commonsense reasoning}.
\newblock In \emph{Proceedings of the 2020 Conference on Empirical Methods in Natural Language Processing (EMNLP)}, pages 2362--2376, Online. Association for Computational Linguistics.

\bibitem[{Scao et~al.(2022)Scao, Fan, Akiki, Pavlick, Ilic, Hesslow, Castagn{\'{e}}, Luccioni, Yvon, Gall{\'{e}}, Tow, Rush, Biderman, Webson, Ammanamanchi, Wang, Sagot, Muennighoff, del Moral, Ruwase, Bawden, Bekman, McMillan{-}Major, Beltagy, Nguyen, Saulnier, Tan, Suarez, Sanh, Lauren{\c{c}}on, Jernite, Launay, Mitchell, Raffel, Gokaslan, Simhi, Soroa, Aji, Alfassy, Rogers, Nitzav, Xu, Mou, Emezue, Klamm, Leong, van Strien, Adelani, and et~al.}]{workshopBLOOM176BParameterOpenAccess2023}
Teven~Le Scao, Angela Fan, Christopher Akiki, Ellie Pavlick, Suzana Ilic, Daniel Hesslow, Roman Castagn{\'{e}}, Alexandra~Sasha Luccioni, Fran{\c{c}}ois Yvon, Matthias Gall{\'{e}}, Jonathan Tow, Alexander~M. Rush, Stella Biderman, Albert Webson, Pawan~Sasanka Ammanamanchi, Thomas Wang, Beno{\^{\i}}t Sagot, Niklas Muennighoff, Albert~Villanova del Moral, Olatunji Ruwase, Rachel Bawden, Stas Bekman, Angelina McMillan{-}Major, Iz~Beltagy, Huu Nguyen, Lucile Saulnier, Samson Tan, Pedro~Ortiz Suarez, Victor Sanh, Hugo Lauren{\c{c}}on, Yacine Jernite, Julien Launay, Margaret Mitchell, Colin Raffel, Aaron Gokaslan, Adi Simhi, Aitor Soroa, Alham~Fikri Aji, Amit Alfassy, Anna Rogers, Ariel~Kreisberg Nitzav, Canwen Xu, Chenghao Mou, Chris Emezue, Christopher Klamm, Colin Leong, Daniel van Strien, David~Ifeoluwa Adelani, and et~al. 2022.
\newblock \href {https://doi.org/10.48550/ARXIV.2211.05100} {{BLOOM:} {A} 176b-parameter open-access multilingual language model}.
\newblock \emph{CoRR}, abs/2211.05100.

\bibitem[{Singh et~al.(2024)Singh, Vargus, D'souza, Karlsson, Mahendiran, Ko, Shandilya, Patel, Mataciunas, O'Mahony, Zhang, Hettiarachchi, Wilson, Machado, Moura, Krzeminski, Fadaei, Erg{\"{u}}n, Okoh, Alaagib, Mudannayake, Alyafeai, Vu, Ruder, Guthikonda, Alghamdi, Gehrmann, Muennighoff, Bartolo, Kreutzer, {\"{U}}st{\"{u}}n, Fadaee, and Hooker}]{singhAyaDatasetOpenAccess2024}
Shivalika Singh, Freddie Vargus, Daniel D'souza, B{\"{o}}rje~F. Karlsson, Abinaya Mahendiran, Wei{-}Yin Ko, Herumb Shandilya, Jay Patel, Deividas Mataciunas, Laura O'Mahony, Mike Zhang, Ramith Hettiarachchi, Joseph Wilson, Marina Machado, Luisa~Souza Moura, Dominik Krzeminski, Hakimeh Fadaei, Irem Erg{\"{u}}n, Ifeoma Okoh, Aisha Alaagib, Oshan Mudannayake, Zaid Alyafeai, Minh~Chien Vu, Sebastian Ruder, Surya Guthikonda, Emad~A. Alghamdi, Sebastian Gehrmann, Niklas Muennighoff, Max Bartolo, Julia Kreutzer, Ahmet {\"{U}}st{\"{u}}n, Marzieh Fadaee, and Sara Hooker. 2024.
\newblock \href {https://doi.org/10.48550/ARXIV.2402.06619} {Aya dataset: An open-access collection for multilingual instruction tuning}.
\newblock \emph{CoRR}, abs/2402.06619.

\bibitem[{Tikhonov and Ryabinin(2021)}]{tikhonov-ryabinin-2021-heads}
Alexey Tikhonov and Max Ryabinin. 2021.
\newblock \href {https://doi.org/10.18653/v1/2021.findings-acl.310} {{I}t{'}s {A}ll in the {H}eads: {U}sing {A}ttention {H}eads as a {B}aseline for {C}ross-{L}ingual {T}ransfer in {C}ommonsense {R}easoning}.
\newblock In \emph{Findings of the Association for Computational Linguistics: ACL-IJCNLP 2021}, pages 3534--3546, Online. Association for Computational Linguistics.

\bibitem[{Touvron et~al.(2023)Touvron, Martin, Stone, Albert, Almahairi, Babaei, Bashlykov, Batra, Bhargava, Bhosale, Bikel, Blecher, Canton{-}Ferrer, Chen, Cucurull, Esiobu, Fernandes, Fu, Fu, Fuller, Gao, Goswami, Goyal, Hartshorn, Hosseini, Hou, Inan, Kardas, Kerkez, Khabsa, Kloumann, Korenev, Koura, Lachaux, Lavril, Lee, Liskovich, Lu, Mao, Martinet, Mihaylov, Mishra, Molybog, Nie, Poulton, Reizenstein, Rungta, Saladi, Schelten, Silva, Smith, Subramanian, Tan, Tang, Taylor, Williams, Kuan, Xu, Yan, Zarov, Zhang, Fan, Kambadur, Narang, Rodriguez, Stojnic, Edunov, and Scialom}]{touvronLlama2OpenFoundation2023}
Hugo Touvron, Louis Martin, Kevin Stone, Peter Albert, Amjad Almahairi, Yasmine Babaei, Nikolay Bashlykov, Soumya Batra, Prajjwal Bhargava, Shruti Bhosale, Dan Bikel, Lukas Blecher, Cristian Canton{-}Ferrer, Moya Chen, Guillem Cucurull, David Esiobu, Jude Fernandes, Jeremy Fu, Wenyin Fu, Brian Fuller, Cynthia Gao, Vedanuj Goswami, Naman Goyal, Anthony Hartshorn, Saghar Hosseini, Rui Hou, Hakan Inan, Marcin Kardas, Viktor Kerkez, Madian Khabsa, Isabel Kloumann, Artem Korenev, Punit~Singh Koura, Marie{-}Anne Lachaux, Thibaut Lavril, Jenya Lee, Diana Liskovich, Yinghai Lu, Yuning Mao, Xavier Martinet, Todor Mihaylov, Pushkar Mishra, Igor Molybog, Yixin Nie, Andrew Poulton, Jeremy Reizenstein, Rashi Rungta, Kalyan Saladi, Alan Schelten, Ruan Silva, Eric~Michael Smith, Ranjan Subramanian, Xiaoqing~Ellen Tan, Binh Tang, Ross Taylor, Adina Williams, Jian~Xiang Kuan, Puxin Xu, Zheng Yan, Iliyan Zarov, Yuchen Zhang, Angela Fan, Melanie Kambadur, Sharan Narang, Aur{\'{e}}lien Rodriguez, Robert Stojnic, Sergey Edunov,
  and Thomas Scialom. 2023.
\newblock \href {https://doi.org/10.48550/ARXIV.2307.09288} {Llama 2: Open foundation and fine-tuned chat models}.
\newblock \emph{CoRR}, abs/2307.09288.

\bibitem[{Uhlig et~al.(2024)Uhlig, Schacht, and Barkur}]{uhligPHOENIXOpenSourceLanguage2024}
Matthias Uhlig, Sigurd Schacht, and Sudarshan~Kamath Barkur. 2024.
\newblock \href {http://arxiv.org/abs/2401.10580} {Phoenix: Open-source language adaption for direct preference optimization}.

\bibitem[{{\"{U}}st{\"{u}}n et~al.(2024){\"{U}}st{\"{u}}n, Aryabumi, Yong, Ko, D'souza, Onilude, Bhandari, Singh, Ooi, Kayid, Vargus, Blunsom, Longpre, Muennighoff, Fadaee, Kreutzer, and Hooker}]{ustunAyaModelInstruction}
Ahmet {\"{U}}st{\"{u}}n, Viraat Aryabumi, Zheng~Xin Yong, Wei{-}Yin Ko, Daniel D'souza, Gbemileke Onilude, Neel Bhandari, Shivalika Singh, Hui{-}Lee Ooi, Amr Kayid, Freddie Vargus, Phil Blunsom, Shayne Longpre, Niklas Muennighoff, Marzieh Fadaee, Julia Kreutzer, and Sara Hooker. 2024.
\newblock \href {https://doi.org/10.48550/ARXIV.2402.07827} {Aya model: An instruction finetuned open-access multilingual language model}.
\newblock \emph{CoRR}, abs/2402.07827.

\bibitem[{Wang et~al.(2023)Wang, Kordi, Mishra, Liu, Smith, Khashabi, and Hajishirzi}]{wangSelfInstructAligningLanguage2023}
Yizhong Wang, Yeganeh Kordi, Swaroop Mishra, Alisa Liu, Noah~A. Smith, Daniel Khashabi, and Hannaneh Hajishirzi. 2023.
\newblock \href {https://doi.org/10.18653/V1/2023.ACL-LONG.754} {Self-instruct: Aligning language models with self-generated instructions}.
\newblock In \emph{Proceedings of the 61st Annual Meeting of the Association for Computational Linguistics (Volume 1: Long Papers), {ACL} 2023, Toronto, Canada, July 9-14, 2023}, pages 13484--13508. Association for Computational Linguistics.

\bibitem[{Ye et~al.(2023)Ye, Kim, Kim, Hwang, Kim, Jo, Thorne, Kim, and Seo}]{yeFLASKFinegrainedLanguage2023a}
Seonghyeon Ye, Doyoung Kim, Sungdong Kim, Hyeonbin Hwang, Seungone Kim, Yongrae Jo, James Thorne, Juho Kim, and Minjoon Seo. 2023.
\newblock \href {https://doi.org/10.48550/ARXIV.2307.10928} {{FLASK:} fine-grained language model evaluation based on alignment skill sets}.
\newblock \emph{CoRR}, abs/2307.10928.

\bibitem[{Yulianto and Supriatnaningsih(2021)}]{yulianto2021google}
Ahmad Yulianto and Rina Supriatnaningsih. 2021.
\newblock Google translate vs. deepl: a quantitative evaluation of close-language pair translation (french to english).
\newblock \emph{AJELP: Asian Journal of English Language and Pedagogy}, 9(2):109--127.

\bibitem[{Zhang et~al.(2023)Zhang, Dong, Li, Zhang, Sun, Wang, Li, Hu, Zhang, Wu, and Wang}]{zhangInstructionTuningLargeSurvey2023}
Shengyu Zhang, Linfeng Dong, Xiaoya Li, Sen Zhang, Xiaofei Sun, Shuhe Wang, Jiwei Li, Runyi Hu, Tianwei Zhang, Fei Wu, and Guoyin Wang. 2023.
\newblock \href {https://doi.org/10.48550/ARXIV.2308.10792} {Instruction tuning for large language models: {A} survey}.
\newblock \emph{CoRR}, abs/2308.10792.

\bibitem[{Zheng et~al.(2023)Zheng, Chiang, Sheng, Zhuang, Wu, Zhuang, Lin, Li, Li, Xing, Zhang, Gonzalez, and Stoica}]{zhengJudgingLLMasajudgeMTBench2023}
Lianmin Zheng, Wei{-}Lin Chiang, Ying Sheng, Siyuan Zhuang, Zhanghao Wu, Yonghao Zhuang, Zi~Lin, Zhuohan Li, Dacheng Li, Eric~P. Xing, Hao Zhang, Joseph~E. Gonzalez, and Ion Stoica. 2023.
\newblock \href {https://doi.org/10.48550/ARXIV.2306.05685} {Judging llm-as-a-judge with mt-bench and chatbot arena}.
\newblock \emph{CoRR}, abs/2306.05685.

\bibitem[{Zhou et~al.(2023)Zhou, Liu, Xu, Iyer, Sun, Mao, Ma, Efrat, Yu, Yu, Zhang, Ghosh, Lewis, Zettlemoyer, and Levy}]{zhouLIMALessMore2023}
Chunting Zhou, Pengfei Liu, Puxin Xu, Srini Iyer, Jiao Sun, Yuning Mao, Xuezhe Ma, Avia Efrat, Ping Yu, Lili Yu, Susan Zhang, Gargi Ghosh, Mike Lewis, Luke Zettlemoyer, and Omer Levy. 2023.
\newblock \href {https://doi.org/10.48550/ARXIV.2305.11206} {{LIMA:} less is more for alignment}.
\newblock \emph{CoRR}, abs/2305.11206.

\end{thebibliography}

\appendix
\clearpage

\section{Preliminary Dataset Study}
\label{sec:dataset_study}
For selecting appropriate multilingual datasets to instruction-tune the multilingual models, we conducted a preliminary study to investigate the language distribution and length statistics. The results, depicted in Table~\ref{tab:dataset_stats} show a high skew towards English-focused datasets.

\section{LIMA Validation Dataset Creation}
\label{appendix:lima-val}
We utilize the open-source project \textit{stackexchange-dataset}\footnote{\url{https://github.com/EleutherAI/stackexchange-dataset}} to scrape 18 different StackExchange topics (bioinformatics, biology, buddhism, chemistry, coffee, cooking, crafts, datascience, diy, earthscience, fitness, health, math, mythology, physics, poker, softwareengineering, travel). 
Despite already filtering for quality by stackexchanges scoring method 
, we end up with 62021 \gls{QA} pairs.
Additionally, we filter answers by phrases such as "my", "as mentioned", "stack exchange", "referenced", "figure", "image", among others, to exclude examples not written in the style of a helpful assistant or referencing images, which cannot be represented in our unimodal models.
We also filter by the length of \gls{QA} pairs, i.e., only allowing pairs that count more words than 1200 but do not exceed 4096 words.
Additionally, we filter by consistent language across question and answer and perform near deduplication with Shingling, MinHashing, and LSH over the \textit{LIMA} training dataset split.
In total, we reduced the 62021 examples to only 84, which we then carefully inspected and manually curated by rewriting or deleting samples. This leads to the final 52 samples, which is roughly the size of the validation dataset reported by \citet{zhouLIMALessMore2023}. 

\section{Translation Quality}
\label{appendix:translation-problems}
Most similar to our benchmark translation efforts is the dataset \textit{MT-Bench-TrueGerman}.
The authors of \textit{MT-Bench-TrueGerman} \footnote{\url{https://huggingface.co/VAGOsolutions/SauerkrautLM-7b-v1}}, report translation problems when using \texttt{GPT-4} as a translation engine. To assess the translation quality of \textit{MT-Bench-X}, we compare their findings with our translations by DeepL.
While \texttt{GPT-4} can translate across various languages, it falls short compared to specialized translation engines such as DeepL.
We showcase this in Table \ref{tab:MT-Bench-translate}, by comparing the failure cases reported by \textit{MT-Bench-TrueGerman} authors.

DeepL offers a more realistic translation than GPT-4 for the anglicism problem and we find the translation of simile accurate.
With the exception of the translation errors due to intentionally grammatically incorrect sources, we cannot support the findings of \textit{MT-Bench-TrueGerman}.

\section{Multilingual Pre-Trained Models}
\label{appendix:model}
To investigate multilingual instruction-tuning performance, we require the pre-trained model to have been i.) trained on multilingual data, including our target languages, ii.) trained with a fair tokenizer, i.e., an equal amount of tokens per language was used for the tokenizer training, and iii.) trained on at least 1T tokens to ensure a minimal set of learned aptitudes as a prerequisite for minimal instruction-following capability.
To the best of our knowledge, only three existing, openly available model families are multilingual European ones. This includes BLOOM~\citep{workshopBLOOM176BParameterOpenAccess2023} and Nemotron~\footnote{\url{https://developer.nvidia.com/nemotron-3-8b}} and \textit{Mixtral-8x7B}. However, BLOOM was not pre-trained on German data and only on 366B tokens for 46 languages, and for Nemotron, no details about the tokenizer training nor details about the dataset language composition are available. While for \textit{Mixtral-8x7B} no details about the tokenizer training are known, it shows high performance across languages for downstream evaluation tasks.

This study is based on an intermediate checkpoint of a multilingual pre-trained model that was trained on 1T tokens at the point when the experimentation for this study was conducted and is presented in~\cite{ali2024progressreporteuropeanllms}. The tokenizer was specifically trained to support each of the 24 EU languages equally. The choice of Tokenizer training parameters and implementation are hereby inferred from results of \citet{aliTokenizerChoiceLLM2023}, who recommend for multilingual models an Unigram~\citep{kudoUnigram2018} SentencePiece\footnote{\url{https://github.com/google/sentencepiece}} Tokenizer with a large vocabulary of 250680 tokens.
The pre-training datasets utilized exhibit an English-dominated share of all 24 European languages. Thereby the five languages in question for this study are contained to the following proportions within the pre-training dataset:
43.88\% EN,
8.65\% FR,
7.63\% ES,
8.48\% DE,
4.64\% IT and
4.51\% Code
). The dataset was filtered and deduplicated. 
Regarding architectural choices, the architecture of Llama 2~\citep{touvronLlama2OpenFoundation2023} was followed.
Regarding the multilingual \textit{Mixtral-8x7B-v0.1}, which supports all langauges inspected within this study, we refer to the details provided by Jiang et al.~\cite{jiangMixtralExperts2024}.

\section{Licenses}
\label{appendix:licenses}
The created \textit{LIMA-X} datasets are licensed by CC BY-NC-SA (Lima-X) or stricter as required by \citet{zhouLIMALessMore2023}, and the used \textit{Bactrian-X} dataset was released under CC BY-NC 4.0 are thus both non-commercial, but usable as intended by the source datasets.

We license the created resource \textit{MT-Bench-X} under Apache License 2.0. This is consistent with the intended use of the source dataset \textit{MT-Bench}~\citep{zhengJudgingLLMasajudgeMTBench2023}.

\section{Evaluation}
\label{appendix:mt_bench_single_score_table_radar}

\subsection{Single Score Evaluation}
\label{appendix:mt_bench_single_score_radar}
\begin{figure*}[ht!]
     \centering
    \begin{subfigure}[b]{\textwidth}
         \centering
         \includegraphics[clip, trim=0.7cm 1.2cm 0.25cm 0.1cm, width=\textwidth]{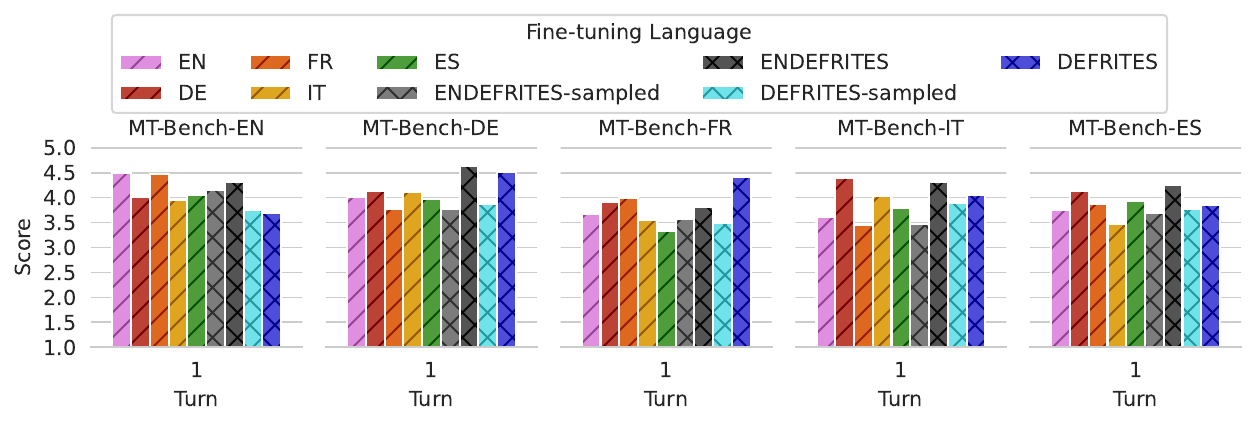}
         \caption{24EU-7B - Bactrian-X}
     \end{subfigure}
     \begin{subfigure}[b]{\textwidth}
         \centering
         \includegraphics[clip, trim=0.7cm 0.6cm 0.2cm 0.9cm, width=\textwidth]{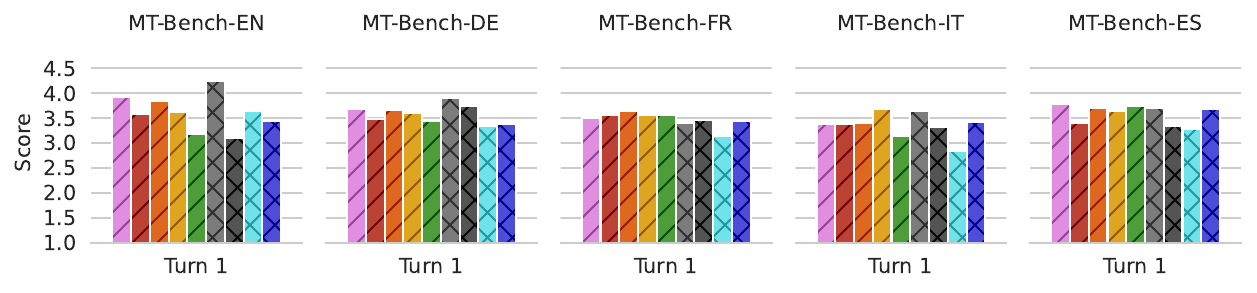}
         \caption{24EU-7B - Bactrian-X-small}
     \end{subfigure}
     \begin{subfigure}[b]{\textwidth}
         \centering
         \includegraphics[clip, trim=0.7cm 0.6cm 0.2cm 1.7cm, width=\textwidth]{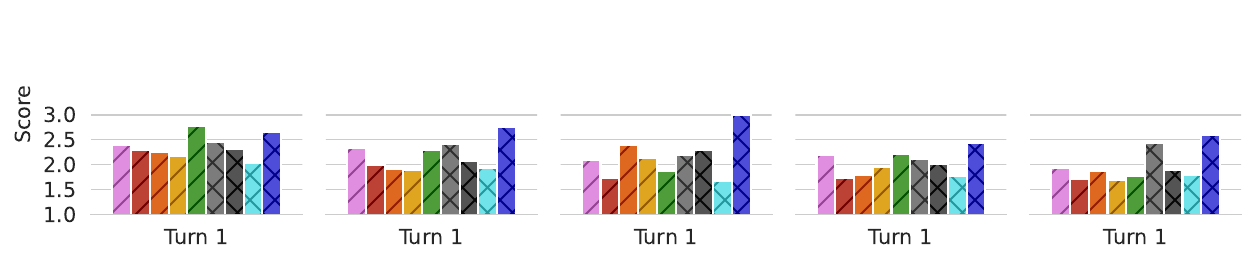}
         \caption{24EU-7B - Lima-X}
     \end{subfigure}
     \begin{subfigure}[b]{\textwidth}
         \centering
         \includegraphics[clip, trim=0.7cm 0.6cm 0.2cm 1cm, width=\textwidth]{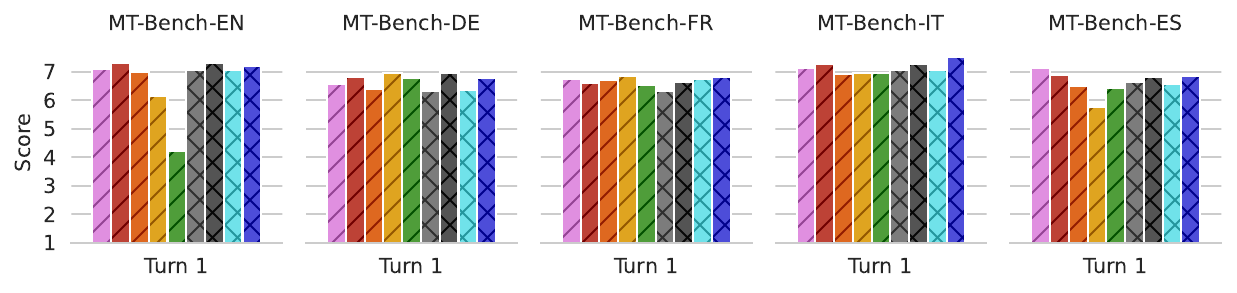}
         \caption{Mixtral-8x7B - Bactrian-X-small}
     \end{subfigure}
     \begin{subfigure}[b]{\textwidth}
         \centering
         \includegraphics[clip, trim=0.7cm 0.25cm 0.2cm 0.2cm, width=\textwidth]{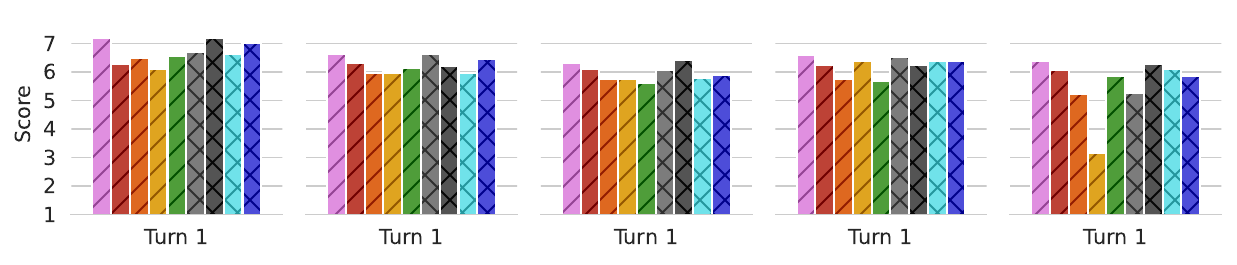}
         \caption{Mixtral-8x7B - Lima-X}
     \end{subfigure}
     \caption{GPT-4-as-a-judge single evaluation average scores for each language mix dataset variant on MT-Bench-X.}
     \label{fig:mt-bench-x-catplot}
\end{figure*}

With the bar plots in Figure~\ref{fig:mt-bench-x-catplot}, we compare differences of monolingual and multilingual instruction-tuned models across all languages within \textit{MT-Bench-X} for all pre-trained models instruction-tuned on all dataset variants. For the main discussion of these results, we refer the reader to the monolingual performance analysis in Section~\ref{sec:single_score_eval}.

\begin{figure*}
     \centering
     \begin{subfigure}[b]{0.495\textwidth}
         \centering
         \includegraphics[clip, trim=0 0 0.5cm 0, width=\textwidth]{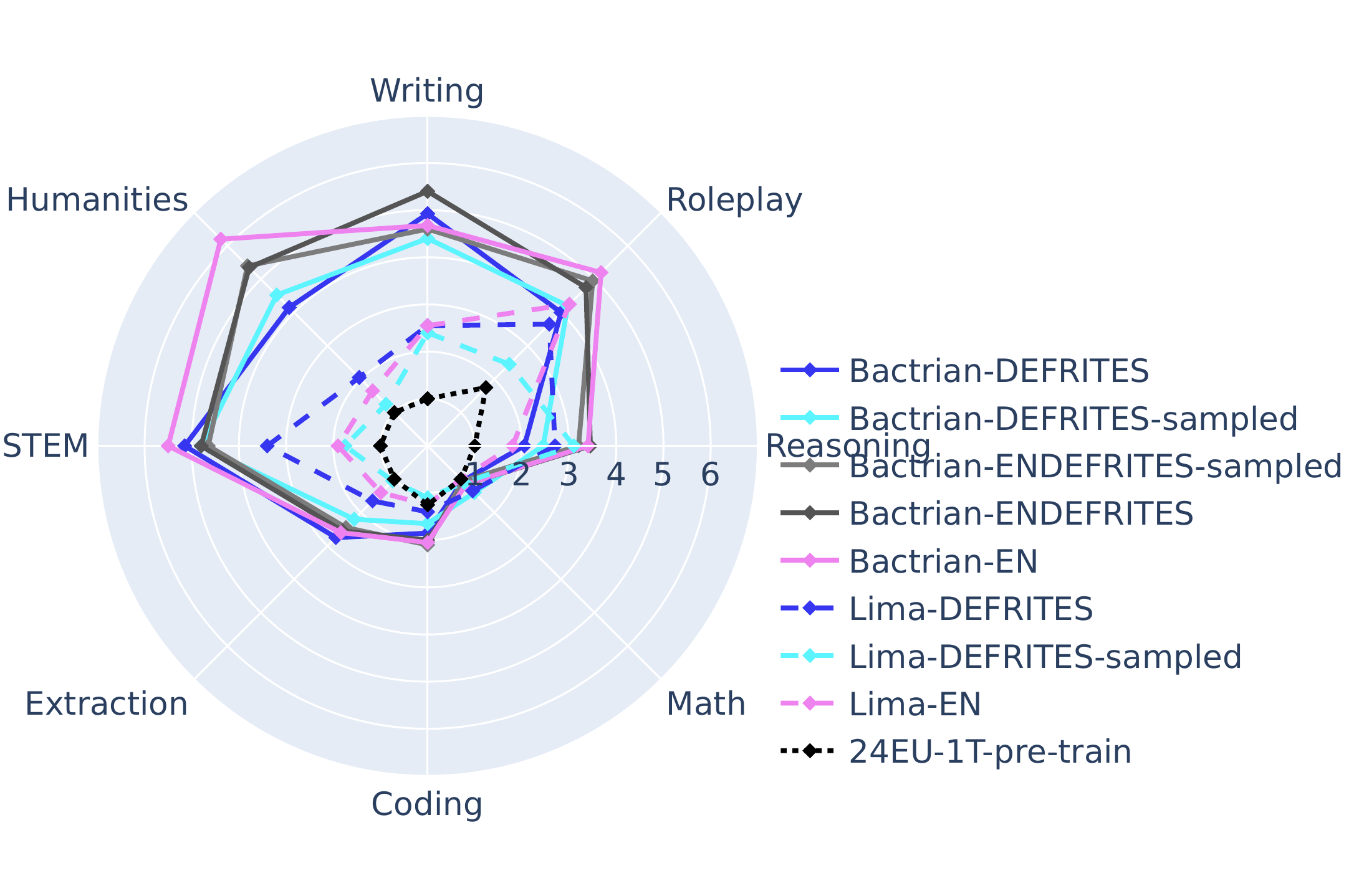}
        \caption{Category-wise MT-Bench-EN single-score by GPT-4.}
        \label{fig:radar-gpt-4-mt-bench-EN}
    \end{subfigure}
    \hfill
    \begin{subfigure}[b]{0.495\textwidth}
         \centering
         \includegraphics[clip, trim=0 0 0.5cm 0, width=\textwidth]{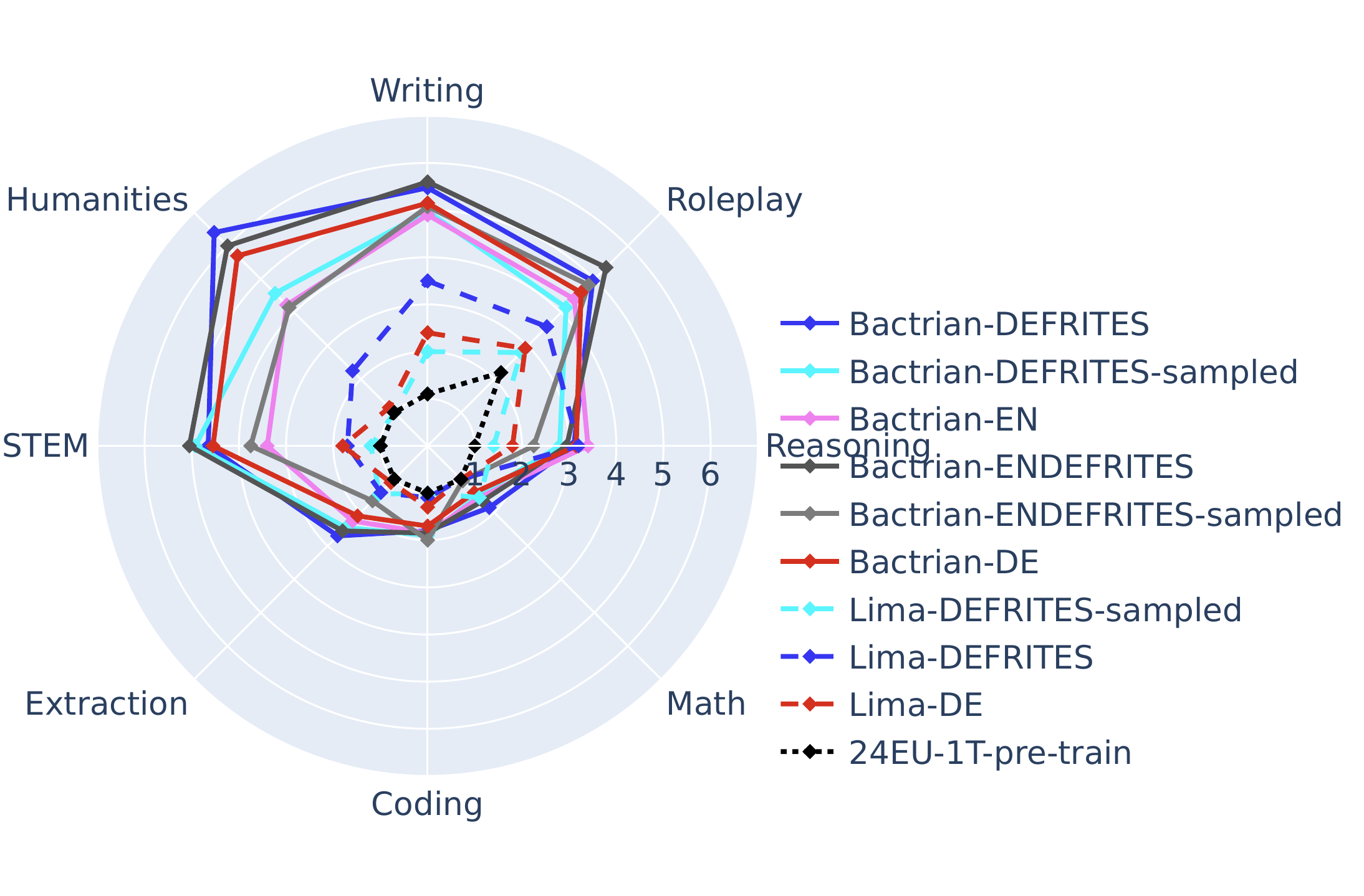}
        \caption{Category-wise MT-Bench-DE single-score by GPT-4.}
        \label{fig:radar-gpt-4-mt-bench-DE}
    \end{subfigure}    
    \begin{subfigure}[b]{0.495\textwidth}
         \centering
         \includegraphics[clip, trim=0 0 0.5cm 0, width=\textwidth]{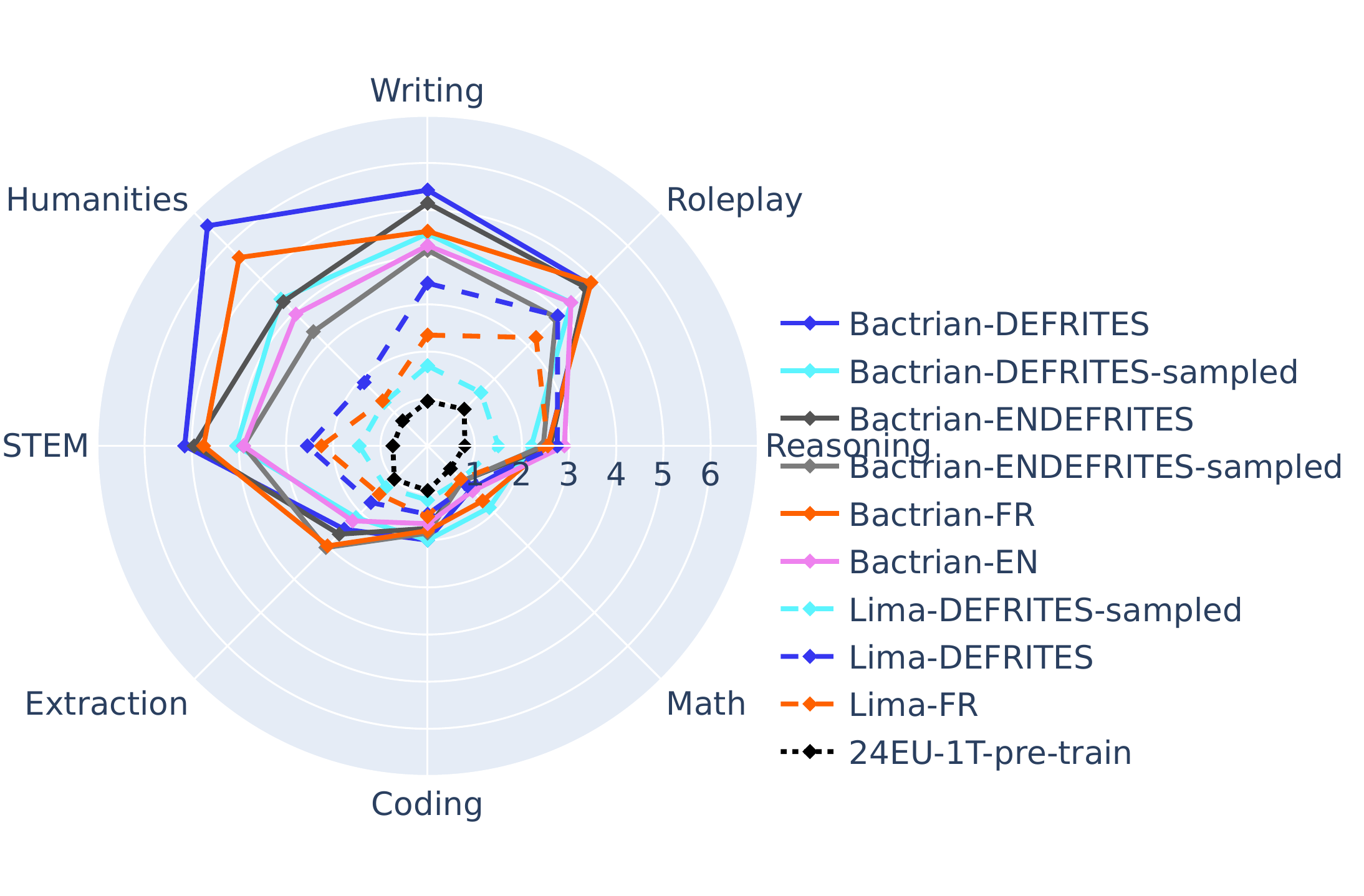}
        \caption{Category-wise MT-Bench-FR single-score by GPT-4.}
        \label{fig:radar-gpt-4-mt-bench-FR}
    \end{subfigure}
    \hfill
    \begin{subfigure}[b]{0.495\textwidth}
         \centering
         \includegraphics[clip, trim=0 0 0.5cm 0, width=\textwidth]{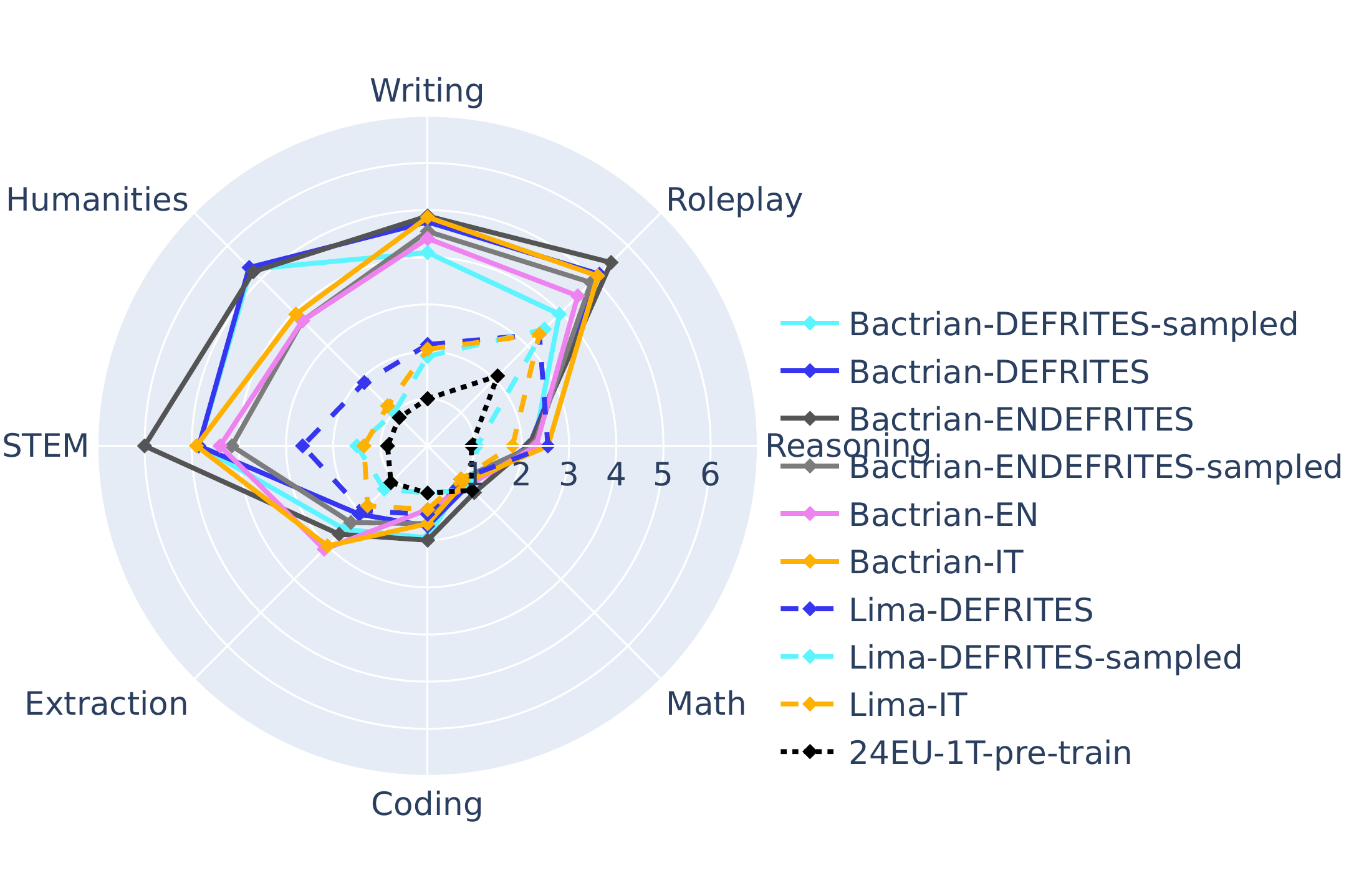}
        \caption{Category-wise MT-Bench-IT single-score by GPT-4.}
        \label{fig:radar-gpt-4-mt-bench-IT}
    \end{subfigure}    
    \begin{subfigure}[b]{0.495\textwidth}
         \centering
         \includegraphics[clip, trim=0 0 0.5cm 0, width=\textwidth]{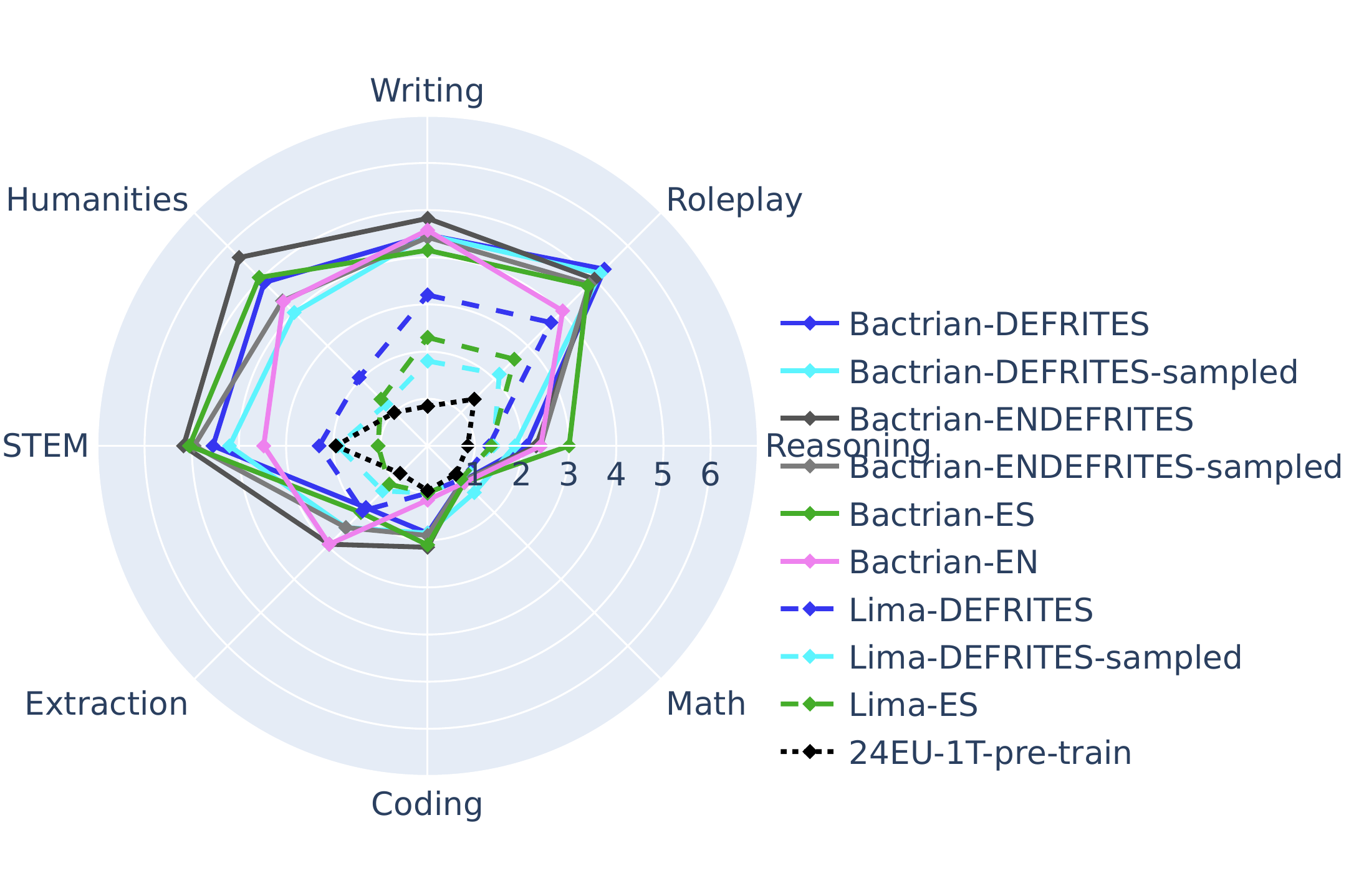}
        \caption{Category-wise MT-Bench-ES single-score by GPT-4.}
        \label{fig:radar-gpt-4-mt-bench-ES}
    \end{subfigure}
    \caption{In-depth MT-Bench-X quality assessment by GPT-4.}
    \label{fig:radar-gpt-4-mt-bench-X}
\end{figure*}

Furthermore, with the radar plots in Figure~\ref{fig:radar-gpt-4-mt-bench-X}, we compare differences of monolingual and multilingual instruction-tuned models across all languages and instruction categories within \textit{MT-Bench-X} for \textit{Bactrian-X} and \textit{Lima-X}. \textit{GPT-4-as-a-judge} was utilized to rate the model responses for each individually and turn-based on a range from 1 to 10.
The highest observed score for the pre-trained \textit{24EU-7B} model is 2.2, which renders its instruction-following capabilities insufficient across languages. While scores increase marginally, the \textit{24EU-7B} base model instruction-tuned on \texttt{Lima-X} shows to perform only on the lower end of the scale. However, we note spikes in the category \textit{Writing} and \textit{Roleplay} across languages.
For \texttt{Bactrian-X}-based \textit{24EU-7B} models, a pattern is evident, showing the best performance for \texttt{Bactrian-ENDEFRITES} or models trained directly in the language of the benchmark. However, here no consistent superior fine-tuning is evident across categories and languages.

\begin{figure*}[ht!]
    \begin{subfigure}[b]{\textwidth}
         \centering
         \includegraphics[clip, trim=1.3cm 0.2cm 1.2cm 0.1cm, width=\textwidth]{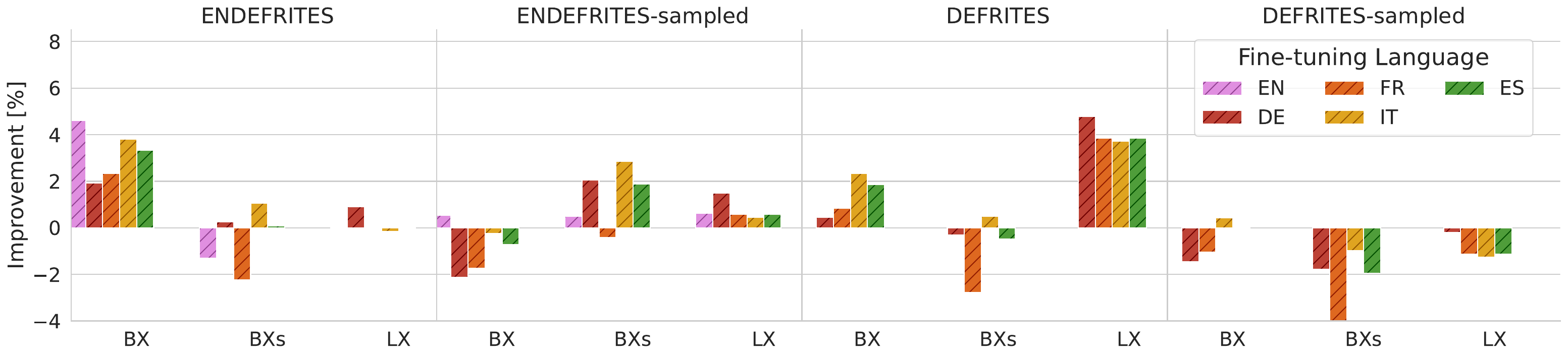}
         \caption{24EU-7B Model}
     \end{subfigure}
     \begin{subfigure}[b]{\textwidth}
         \centering
         \includegraphics[clip, trim=1.3cm 0.2cm 1.2cm 0.1cm, width=\textwidth]{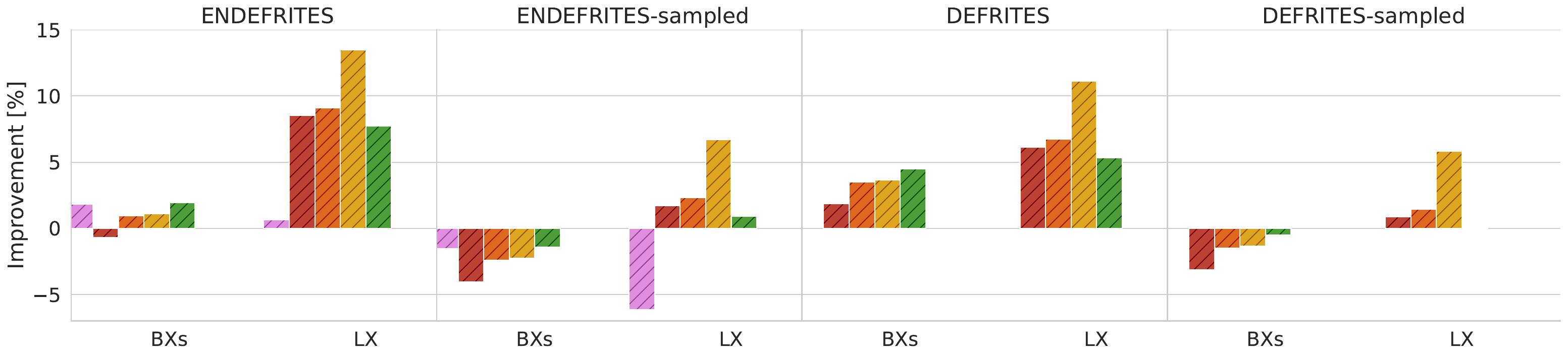}
         \caption{Mixtral-8x7B Model}
     \end{subfigure}
    \caption{Percentage improvement for turn two averaged across \textit{MT-Bench-X} languages of models fine-tuned on parallel mixed language instruction-tuning datasets over single language fine-tunings.}
    \label{fig:avg-mt-bench-x-perc-improv-turn-two}
\end{figure*}

\begin{figure*}
     \centering
     \begin{subfigure}[b]{0.495\textwidth}
         \centering
         \includegraphics[clip, trim=0cm 0.7cm 0cm 0cm, width=\textwidth]{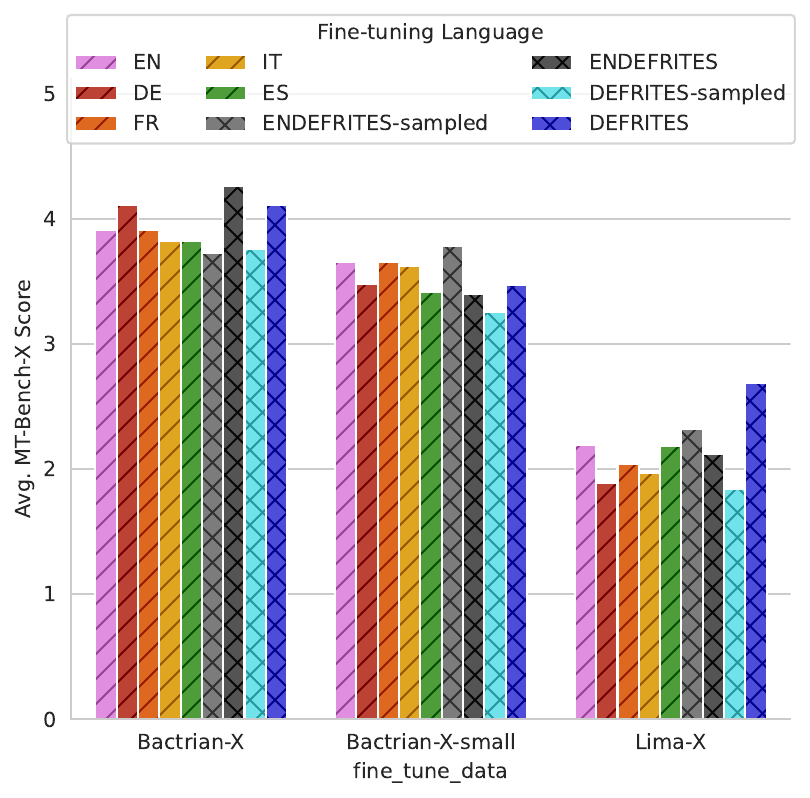}
        \caption{24EU-7B}
        \label{fig:catplot-cross-lingual-absolute-24EU-7B}
    \end{subfigure}
    \hfill
    \begin{subfigure}[b]{0.495\textwidth}
         \centering
         \includegraphics[clip, trim=0cm 0.7cm 0.4cm 0, width=\textwidth]{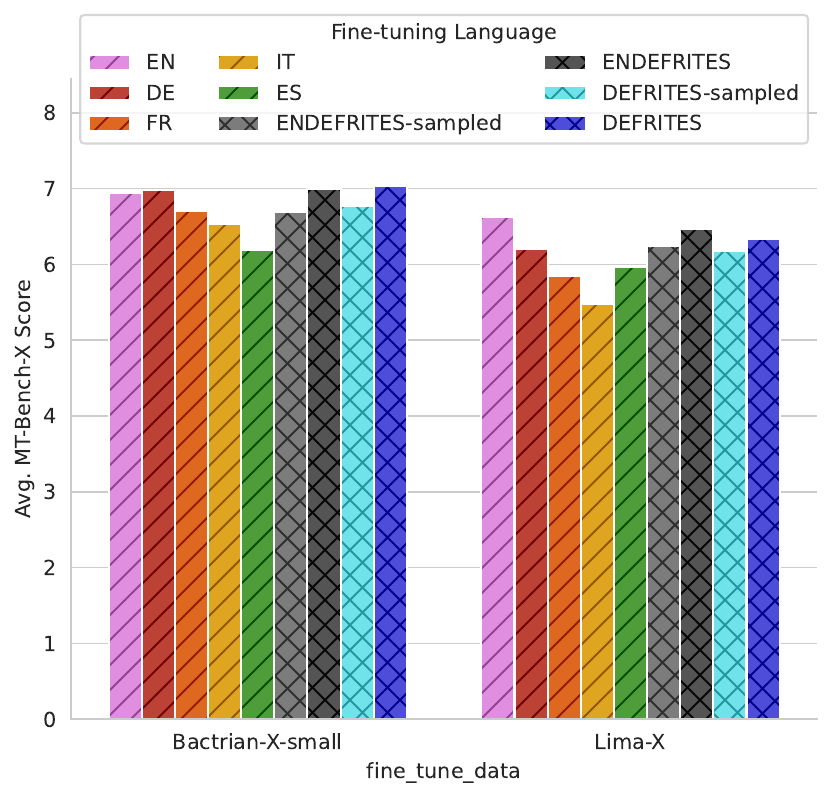}
        \caption{Mixtral-8x7B}
        \label{fig:catplot-cross-lingual-absolute-mixtral}
    \end{subfigure}    
    \caption{Absolute cross-lingual MT-Bench-X scores across all five languages for turn one.}
    \label{fig:avg-mt-bench-x-abs-scores-turn-one}
\end{figure*}

For completeness, we display the evaluation results of the percentage average performance improvement of models fine-tuned on parallel mixed language instruction-tuning datasets across \textit{MT-Bench-X} languages for turn two in Figure~\ref{fig:avg-mt-bench-x-perc-improv-turn-two}. The same pattern as for turn one within Figure~\ref{fig:avg-mt-bench-x-perc-improv-turn-1} in Section~\ref{sec:single_score_eval} is evident.
Additionally, we report the absolute cross-lingual MT-Bench-X scores by averaging across the evaluation results for all five languages for turn one in Figure \ref{fig:avg-mt-bench-x-abs-scores-turn-one}.

\paragraph{Detailed Tabular View}
\label{appendix:single-mt-bench-x-tables}
In the Tables \ref{tab:GPT-4-single-score-Mt-bench-ENDE}, \ref{tab:GPT-4-single-score-Mt-bench-FRIT} and \ref{tab:GPT-4-single-score-Mt-bench-ES}, the detailed category-wise scores of each model variant are displayed. Despite being an automatic evaluation and less costly, still high costs are involved in utilizing the OpenAI API for evaluation with \textit{GPT-4-as-judge}. Because of these high costs involved, we did not average the results across multiple evaluation runs.

\subsection{Pair-wise Evaluation}
\label{appendix:pair_wise_eval_prompts}
For the pair-wise mode of evaluating according to \textit{MT-Bench}~\citep{zhengJudgingLLMasajudgeMTBench2023} in German, we translate the English prompts for \textit{GPT-4-as-a-judge} as displayed in Table~\ref{tab:pair-wise-eval-prompts}.

\begin{figure}[h]
         \centering
        \includegraphics[width=\columnwidth]{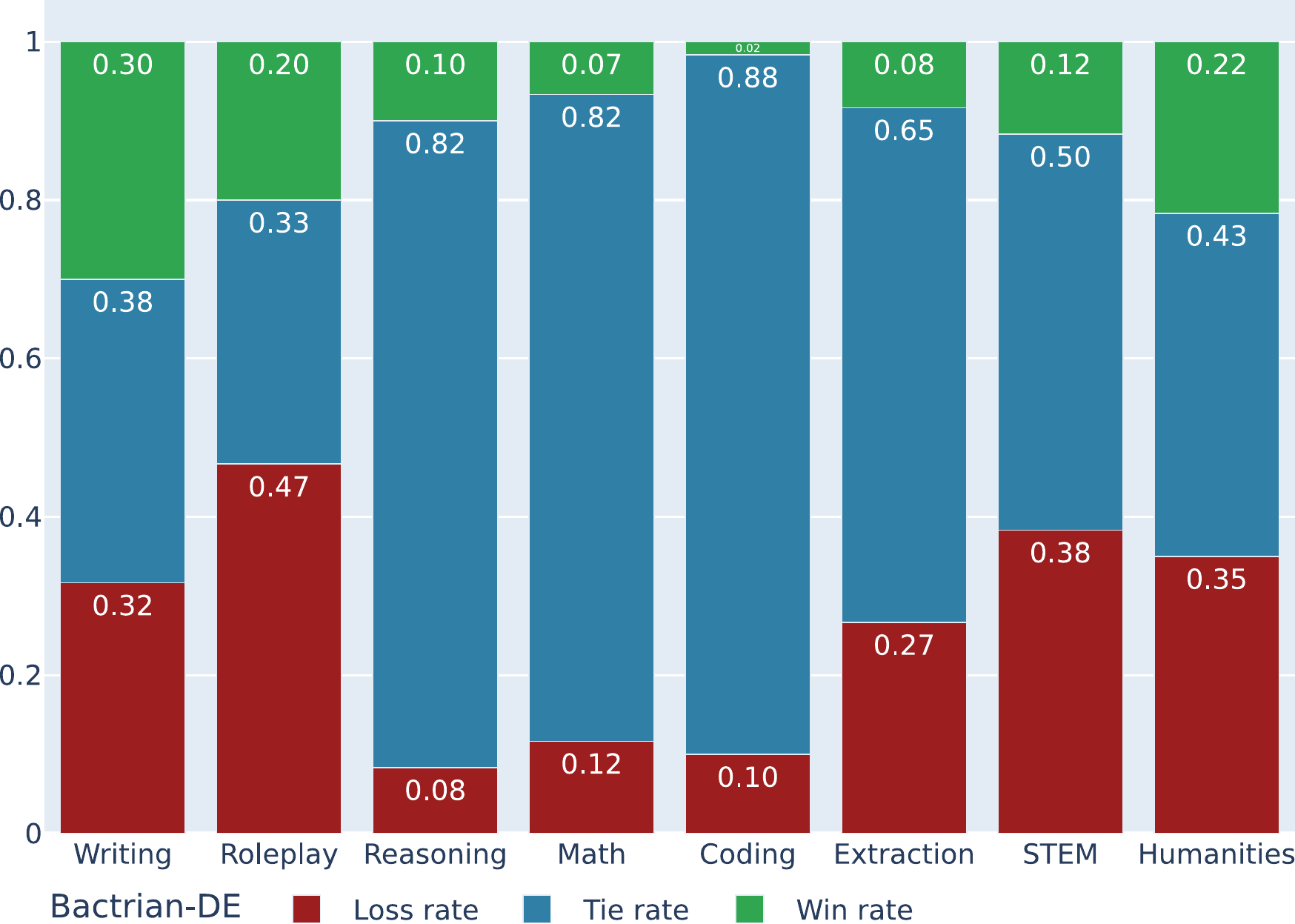}
        \caption{\texttt{Bactrian-DE} vs.  \texttt{Bactrian-ENDEFRITES}. 
        Pair-wise MT-Bench-DE quality assessment by GPT-4.
        }
        \label{fig:pair-wise-gpt-4-mt-bench-DE}
\end{figure}

The initial experiment, according to the methods of \citet{zhengJudgingLLMasajudgeMTBench2023}, involved only the three voting options \textit{Model A is better}, \textit{Model B is better} and \textit{Tie}. However, as evident from Figure~\ref{fig:pair-wise-gpt-4-mt-bench-DE}, an imbalance between the voting options occurs, and it is unclear whether the large fraction of \textit{Tie} votes result from equally bad or good answers.
To clarify whether the large fraction of \textit{Tie} votes occurs due to insufficient overall performance, we repeated the experiment with the additional option to select when both models are equally bad, as presented in Section~\ref{sec:pair_wise_human_eval}.

\subsection{Human Evaluation}
\label{appendix:human_eval}
\begin{figure*}
     \centering
     \includegraphics[width=\textwidth]{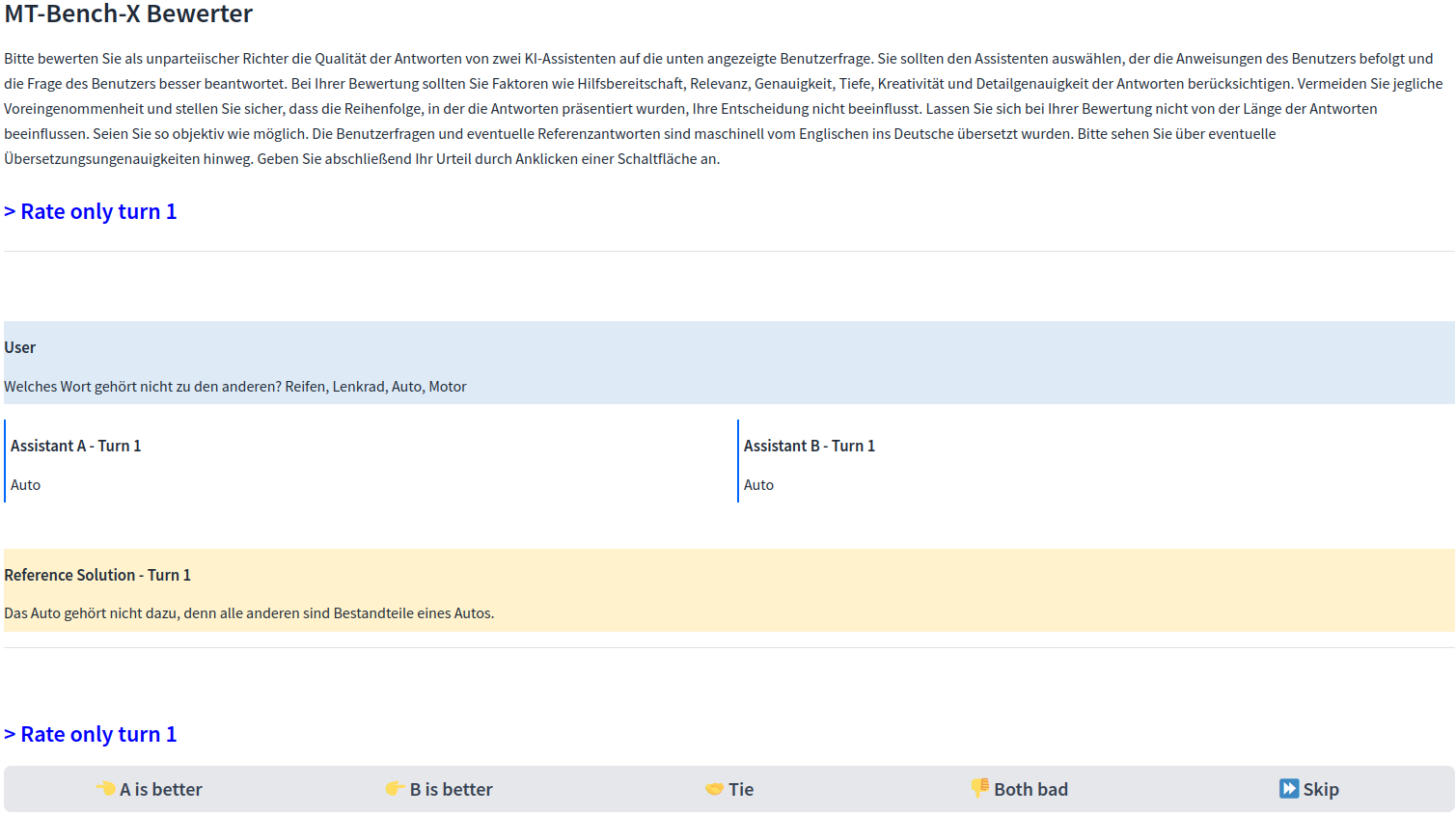}
     \caption{User interface for human evaluation. The model response sides are selected randomly for each turn. Each evaluation instance is sampled randomly and comprises of the two-turn responses of two models, which are evaluated one after the other. Based on~\citet{zhengJudgingLLMasajudgeMTBench2023}.}
     \label{fig:MT_Bench_human_eval}
\end{figure*}

We conduct a human evaluation on pair-wise model assessment on the German \textit{MT-Bench-DE}. Therefore, we provide the annotation platform depicted in Figure~\ref{fig:MT_Bench_human_eval}, in which academic, native-speaking Germans pair-wise assess model responses by selecting either a clear winner, denoting a tie or signaling that both models were not helpful at all in answering the multi-turn user request. The 12 participants are all considered to be experts, as all are working within the field of generative artificial intelligence. All participants were informed that the purpose of the annotations was to investigate research questions for a paper to be published. No payment was involved. Consent for the human evaluation study as well as the manual curation of \textit{MT-Bench-X} was obtained through video calls. 
To protect the annotators participating in this study, we anonymized the data collected.
The translation for the user directives in English is as follows: 
\begin{quote}
"\textit{MT-Bench-X Evaluator
Please act as an impartial judge and evaluate the quality of the responses provided by two AI assistants to the user question displayed below. You should choose the assistant that follows the user's instructions and answers the user's question better. 
Your evaluation should consider factors such as the helpfulness, relevance, accuracy, depth, creativity, and level of detail of their responses. 
Avoid any positional biases and ensure that the order in which the responses were presented does not influence your decision. 
Do not allow the length of the responses to influence your evaluation. 
Do not favor certain names of the assistants. 
Be as objective as possible. 
Finally, indicate your verdict by clicking one button. 
Please use the tie button sparsely.}" Cf.~\citep{zhengJudgingLLMasajudgeMTBench2023}.
\end{quote}

\section{Qualitative Analysis}
\label{appendix:qaulitative_analysis}
In the following sections, we conduct a qualitative analysis regarding cross-lingual vs. monolingual model performance (Cf. Appendix~\ref{appendix:qual_cross_mono}) and the performance of GPT-4 as a judge (Cf. Appendix~\ref{appendix:qual_gpt4-as-a-judge}).

\subsection{Cross-lingual vs. Monolingual}
\label{appendix:qual_cross_mono}
Within the examples shown in Table~\ref{tab:qual_analysis} and Table~\ref{tab:qual_analysis2}, minor differences in the format of the answers of \texttt{Bactrian-ENDEFRITES} and \texttt{Bactrian-DE} can be observed.
While the multilingual fine-tuned model shows a format and placeholders as one would expect, the model \texttt{Bactrian-DE} shows incorrect formatting.
We also noted that responses of \texttt{Bactrian-ENDEFRITES} are often more elaborate compared to monolingual \texttt{Bactrian-X} models.

\subsection{GPT-4-as-a-judge}
\label{appendix:qual_gpt4-as-a-judge}
While for both examples, reasonably high ratings are given by \textit{GPT-4-as-a-judge} during single mode evaluation and reasonable justifications for not scoring model outputs higher where given by GPT-4 (not shown), we can observe inconsistencies and incorrect ratings, as marked red in Table~\ref{tab:qual_analysis3} - even when a reference solution is given - for a lemon-picked example.
While this exemplifies that human evaluation should still be considered the gold standard, it also shows that a coarse impression of quantitative model performance by \textit{GPT-4-as-a-judge} is a cost- and time-effective evaluation option despite not being as accurate for some examples.

\section{Infrastructure \& Computational Costs}
We trained each of our 27 7B parameter models and 18 8x7B parameter models on NVIDIA H100 GPUs, and the training and answer generation of the models took between 337 (LIMA-X) and 540 (Bactrian-X) and 337 (Bactrian-X-small) GPU hours.
Therefore, the total training and answer generation costs amounted to $\approx 1214$ GPU hours.

\clearpage

\begin{table*}[!htbp]
    \centering
    \begin{tabular}{llrrrrr>{\hspace{0.5em}}rr}
        \toprule
        & & \multicolumn{5}{c}{\textbf{Samples per Language}} & \multicolumn{2}{c}{\textbf{$\varnothing$ Words}} \\
        
        \textbf{Dataset} & \textbf{Source} & \textbf{EN} & \textbf{DE} & \textbf{FR} & \textbf{IT} & \textbf{ES} & \textbf{User} & \textbf{Sys.}\\ 
        \midrule
                                                                        xP3mt \citep{muennighoffCrosslingualGeneralizationMultitask2023}    & template                 & 32.7M & 0 & 5M & 0 & 5.1M              & 147 & 47\\ 
        HH-RLHF \citep{baiTrainingHelpfulHarmless2022}\tnote{a}                  & gen.             & 112k & 13 & 10 & 8 & 9                     & 118 & 67 \\ 
        OASST \citep{kopfOpenAssistantConversationsDemocratizing2023}\tnote{a}   & gen.             & 20.2k & 1.5k & 1.1k & 359 & 13.4k          & 20 & 114\\ 
        Bactrian-X \citep{liBactrianXMultilingualReplicable2023}            & gen.                 & 67k & 67k & 67k & 67k & 67k                & 27 & 90 \\ 
        LIMA \citep{zhouLIMALessMore2023}\tnote{a}                               & human             & 1k & 4 & 0 & 0 & 2                            & 39 & 419  \\ 
    \bottomrule
    \end{tabular}
    \caption{Language distribution for different instruction datasets for English (EN), German (DE), French (FR), Italian (IT) and Spanish (ES).}
    \label{tab:dataset_stats}
\end{table*}

\begin{table*}[ht!]
    \centering
    \begin{tabularx}{\linewidth}{p{21mm}|XXX}        & \textbf{Original} & \textbf{GPT-4} & \textbf{DeepL} \\
        \textbf{Problem} & \citep{zhengJudgingLLMasajudgeMTBench2023} & \textbf{MT-Bench-TrueGerman} & \textbf{MT-Bench-DE} \\
        \toprule
        Anglicism & Now you are a machine learning engineer \dots & Jetzt bist du ein Maschinenlern-Ingenieur. \dots & Jetzt sind Sie ein Ingenieur für maschinelles Lernen. \dots \\
        \midrule
        Grammatically incorrect source & \dots correct any grammatical errors: She didn't \textit{remembre} \textit{where is her} purse, so I \textit{thinks} its & \dots grammatikalische Fehler zu korrigieren: Sie erinnerte sich nicht daran, wo ihre Geldbörse ist, also denke ich, \dots & \dots grammatikalische Fehler zu korrigieren: Sie erinnerte sich nicht, wo ihre Handtasche war, also dachte ich, \dots \\
        \midrule
        Reference mismatch & Please assume the role of an English translator, \dots Regardless of the language I use, \dots respond \dots in English. & Bitte nehmen Sie die Rolle eines englischen Übersetzers an \dots auf Englisch antworten. & Bitte schlüpfen Sie in die Rolle eines Englisch-Übersetzers \dots auf Englisch antworten.  \\
        \midrule
        None & Can you rephrase your previous answer and incorporate a metaphor or simile in each sentence? & Kannst du deine vorherige Antwort umformulieren und in jedem Satz eine Metapher oder ein Gleichnis einbauen? & Können Sie Ihre vorherige Antwort umformulieren und in jeden Satz eine Metapher oder ein Gleichnis einbauen? \\
        \bottomrule
    \end{tabularx}
    \caption{EN-DE translation comparison by GPT-4 (MT-Bench-True-german) and DeepL (MT-Bench-DE, Ours)}
    \label{tab:MT-Bench-translate}
\end{table*}

\begin{table*}
\centering
    \begin{tabularx}{\linewidth}{XX}    \toprule
    \multicolumn{1}{c}{\textbf{English Original}}&\multicolumn{1}{c}{\textbf{Translated German}}\\
    \multicolumn{2}{c}{\textbf{Instruction}}\\
    \midrule
      Please act as an impartial judge and evaluate the quality of the responses provided by two AI assistants to the user question displayed below. You should choose the assistant that follows the user's instructions and answers the user's question better. Your evaluation should consider factors such as the helpfulness, relevance, accuracy, depth, creativity, and level of detail of their responses. Begin your evaluation by comparing the two responses and provide a short explanation. Avoid any position biases and ensure that the order in which the responses were presented does not influence your decision. Do not allow the length of the responses to influence your evaluation. Do not favor certain names of the assistants. Be as objective as possible. After providing your explanation, output your final verdict by strictly following this format: \text{"[[A]]"} if assistant A is better, \text{"[[B]]"} if assistant B is better, and \text{"[[C]]"} for a tie.&Bitte beurteilen Sie als unparteiischer Richter die Qualität der Antworten von zwei KI-Assistenten auf die unten dargestellte Benutzerfrage. Sie sollten den Assistenten auswählen, der die Anweisungen des Nutzers befolgt und die Frage des Nutzers besser beantwortet. Bei Ihrer Bewertung sollten Sie Faktoren wie Hilfsbereitschaft, Relevanz, Genauigkeit, Tiefe, Kreativität und Detailgenauigkeit der Antworten berücksichtigen. Beginnen Sie Ihre Bewertung mit einem Vergleich der beiden Antworten und geben Sie eine kurze Erklärung ab. Vermeiden Sie jegliche Voreingenommenheit und stellen Sie sicher, dass die Reihenfolge, in der die Antworten präsentiert wurden, keinen Einfluss auf Ihre Entscheidung hat. Lassen Sie sich bei Ihrer Bewertung nicht von der Länge der Antworten beeinflussen. Bevorzugen Sie nicht bestimmte Namen der Assistenten. Seien Sie so objektiv wie möglich. Geben Sie nach Ihrer Erklärung Ihr endgültiges Urteil ab, indem Sie sich strikt an dieses Format halten: \text{"[[A]]"}, wenn Assistent A besser ist, \text{"[[B]]"}, wenn Assistent B besser ist, \text{"[[C]]"} bei Gleichstand und \text{"[[D]]"} wenn beide Assistenten gleich schlecht sind.\\
      \midrule
        \multicolumn{2}{c}{\textbf{Prompt Template}}\\
        \midrule
        \text{[User Question]}&\text{[Frage des Nutzers]}\\\relax
        \text{\{question\}}&\text{\{question\}}\\\\\relax
        \text{[The Start of Assistant A's Answer]}&\text{[Beginn der Antwort von Assistent A]}\\\relax
        \text{\{answer\_a\}}&\text{\{answer\_a\}}\\\relax
        \text{[The End of Assistant A's Answer]}&\text{[Das Ende der Antwort von Assistent A]}\\\\\relax
        \text{[The Start of Assistant B's Answer]}&\text{[Beginn der Antwort von Assistent B]}\\\relax
        \text{\{answer\_b\}}&\text{\{answer\_b\}}\\\relax
        \text{[The End of Assistant B's Answer]}&\text{[Ende der Antwort von Assistent B]}\\
        \bottomrule
    \end{tabularx}
    \caption{English original prompt for pair-wise evaluation (Cf.~\citep{zhengJudgingLLMasajudgeMTBench2023}) and it's translation and extension by the fourth option \text{[[D]]}: \textit{[\dots] and "[[D]]" if both assistants are equally bad.}}
    \label{tab:pair-wise-eval-prompts}
\end{table*}

\begin{table*}[!ht]
    \centering
    \begin{tabularx}{\linewidth}{X}        \toprule
        \textbf{Request (writing)}
        Draft a professional email seeking your supervisor's feedback on the 'Quarterly Financial Report' you prepared. Ask specifically about the data analysis, presentation style, and the clarity of conclusions drawn. Keep the email short and to the point.\\
        \midrule
        \textbf{\texttt{Bactrian-DE}} \textbf{(Rating 9)}\\         Dear [name of supervisor], I am writing to formally request your feedback on my Quarterly Financial Report. The report was prepared in accordance with the required data analysis and presentation style, as per our company's guidelines. I would appreciate if you could provide me with your feedback on the following:\\
        \begin{itemize}
            \item Data analysis: Can you please comment on the accuracy of the data presented in the report?
            \item Presentation style: How do you feel about the visual presentation of the data? Is it easy to understand and visually appealing?
            \item Conclusions drawn: Please provide your insights on the conclusions drawn from the report. Were they clear and easy to understand, or do you have any concerns or suggestions? I appreciate your time and consideration in reviewing my work, and I look forward to your response. Thank you, [Your Name]
        \end{itemize}\\
        \midrule
        \textbf{\texttt{Bactrian-ENDEFRITES}} \textbf{(Rating 10)}\\\relax
        Dear [Supervisor],\\\\
        I hope this email finds you well. I am writing to request your valuable feedback on my Quarterly Financial Report. Please take a moment to review the report and let me know if there are any areas you would like me to improve on.\\\\
        In regards to the data analysis, I have used a combination of quantitative and qualitative methods to analyze the data and present the results in an easy-to-understand manner. The report clearly demonstrates the financial performance of the company during the quarter, its trends, and the implications for the future. I would appreciate any feedback or suggestions you may have on how to further improve the data analysis.\\\\
        In terms of the presentation style, I have included charts, graphs, and tables to provide visual aids to the report. The report is well-organized with a clear heading structure and a concise conclusion section. I would appreciate any comments on the clarity of the conclusions drawn.\\\\
        Overall, I am satisfied with the quality of the report, but I am always eager to learn from your feedback. Thank you for your time.\\\\
        Sincerely,\\\\\relax
        [Your Name] \\\\
        \bottomrule
    \end{tabularx}
    \caption{Qualitative comparison of answers of the 24EU 7B model fine-tuned on \texttt{Bactrian-ENDEFRITES} and \texttt{Bactrian-DE} for \textit{MT-Bench-EN}.}
    \label{tab:qual_analysis}
\end{table*}

\begin{table*}[!ht]
    \centering
    \begin{tabularx}{\linewidth}{X}        \toprule
        \textbf{Request (stem) - Turn 1}
        The city of Vega intends to build a bridge that will span the Vegona River, covering a distance of 1.8 kilometers. The proposed location falls within a seismically active area that has experienced several high-magnitude earthquakes. Given these circumstances, what would be the best approach to constructing the bridge?\\
        \midrule
        \texttt{\textbf{Bactrian-DE}} \textbf{(Rating 7)}                 The best approach to constructing the bridge would be to use a combination of seismic engineering measures and construction techniques that are well-suited to the seismically active area. This would include the use of reinforced concrete and high-strength steel, as well as the use of advanced design and construction methods that are able to withstand the effects of earthquakes. The bridge would also need to be built in a way that allows for easy access for emergency response teams in the event of an earthquake.\\
        \midrule
        \texttt{\textbf{Bactrian-ENDEFRITES}}  \textbf{(Rating 8)} 
                The best approach to constructing the bridge in a seismically active area would be to use advanced engineering techniques and materials that are designed to withstand high seismic activity. This could include using reinforced concrete, high-strength steel, and advanced seismic design principles to ensure the bridge is able to withstand the forces of an earthquake. Additionally, the bridge could be built on a solid foundation, such as a concrete pier, to further increase its stability. It would also be important to carefully monitor the construction process and ensure that the bridge is built to the highest safety standards.\\
        \midrule
        \textbf{Request (stem) - Turn 2}
        What are the key disadvantages or flaws of your solution? Please perform calculations and use numbers to illustrate them.\\
        \midrule
        \texttt{\textbf{Bactrian-DE}}  \textbf{(Rating 5)}                 The key disadvantages or flaws of the proposed solution are as follows:\\
        \begin{itemize}
            \item The cost of the project would be significantly higher than if the bridge were built in a different location.
            \item The construction process would be more complex and time-consuming, as the bridge would need to be built in a seismically active area.
            \item The bridge would be more vulnerable to earthquakes, as it would be built in a seismically active area.
            \item The construction process would be more dangerous for workers, as they would be working in an area that is prone to earthquakes.
        \end{itemize}\\
        \midrule
        \texttt{\textbf{Bactrian-ENDEFRITES}} \textbf{(Rating 6)}
                The key disadvantages of the proposed solution are as follows:
        \begin{enumerate}
            \item High cost: The use of advanced engineering techniques and materials would increase the cost of the bridge significantly, making it unaffordable for the city of Vega.
            \item Long construction time: The construction process would take a long time, as the bridge would need to be built on a solid foundation and carefully monitored during the construction process.
            \item Environmental impact: The construction process would have a significant impact on the environment, as the bridge would require the clearing of trees and other vegetation, which could have negative consequences for the local ecosystem.
            \item Safety concerns: The bridge would be built in a seismically active area, which could increase the risk of an earthquake damaging the bridge or causing injuries to construction workers.
        \end{enumerate}\\
        \bottomrule
    \end{tabularx}
    \caption{Qualitative comparison of answers of the 24EU 7B model fine-tuned on \texttt{Bactrian-ENDEFRITES} and \texttt{Bactrian-DE} for \textit{MT-Bench-EN}.}
    \label{tab:qual_analysis2}
\end{table*}

\begin{table*}[!ht]
    \centering
    \begin{tabularx}{\linewidth}{X}        \toprule
        \textbf{Request (reasoning) - Turn 1}
        David has three sisters. Each of them has one brother. How many brothers does David have?\\
        \textbf{Reference Solution} David has no brother. He is the one brother of his three sisters.\\
        \midrule
        \texttt{\textbf{Bactrian-DE}}         David has two brothers.\\
        \textbf{Judgment} The assistant's answer is incorrect. [\dots]
                Rating: [[2]]\\
        \texttt{\textbf{Bactrian-DEFRITES}}
        David has six brothers.\\
        \textbf{Judgment} The assistant's answer is incorrect. [\dots]
                Rating: [[1]]\\
        \texttt{\textbf{Bactrian-DEFRITES-sampled}}
        David has two brothers.\\
        \textbf{Judgment} The assistant's answer is incorrect. [\dots]
                Rating: [[2]]\\
        \texttt{\textbf{Bactrian-ENDEFRITES}}
        David has one brother.\\
        \textcolor{red}{\textbf{Judgment}  The assistant's answer is correct. [\dots]
                Rating: [[10]]}\\
        \texttt{\textbf{Bactrian-EN}}
        David has one brother.\\
        \textcolor{red}{\textbf{Judgment} The assistant's answer is correct. [\dots]
                Rating: [[10]]}\\
        \texttt{\textbf{Bactrian-ENDEFRITES-sampled}}
        David has one brother.\\
        \textcolor{red}{\textbf{Judgment} The assistant's answer is correct. [\dots]
                Rating: [[10]]}\\
        \midrule
        \textbf{Request (reasoning) - Turn 2}
        If we change the previous question and assume that each sister of David has two brothers, how many brothers would David have?\\
        \textbf{Reference Solution} David has one brother.\\
        \midrule
        \texttt{\textbf{Bactrian-DE}}         David has one brother.\\
        \textcolor{red}{\textbf{Judgment} The assistant's response to the second question is incorrect. [\dots]
                Rating: [[1]]}\\
        \texttt{\textbf{Bactrian-DEFRITES}}
        David has fourteen brothers.\\
        \textbf{Judgment} The assistant's response is incorrect. [\dots]
                Rating: [[1]]\\
        \texttt{\textbf{Bactrian-DEFRITES-sampled}}
        David would have four brothers.\\
        \textbf{Judgment} The assistant's response to the second question is incorrect. [\dots]
                Rating: [[2]]\\
        \texttt{\textbf{Bactrian-ENDEFRITES}}
        David would have two brothers.\\
        \textcolor{red}{\textbf{Judgment} The assistant's answer is correct. [\dots]
                Rating: [[10]]}\\
        \texttt{\textbf{Bactrian-EN}}
        David has two brothers.\\
        \textcolor{red}{\textbf{Judgment} The assistant's answer is correct. [\dots]
                Rating: [[10]]}\\
        \texttt{\textbf{Bactrian-ENDEFRITES-sampled}}
        David has two brothers.\\
        \textcolor{red}{\textbf{Judgment} The assistant's answer is correct. [\dots]
                Rating: [[10]]}\\
        \bottomrule
    \end{tabularx}
    \caption{Qualitative comparison of \texttt{Bactrian-ENDEFRITES}, \texttt{Bactrian-ENDEFRITES-sampled}, \texttt{Bactrian-DEFRITES}, \texttt{Bactrian-DEFRITES-sampled}, \texttt{Bactrian-EN} and \texttt{Bactrian-DE} on \textit{MT-Bench-EN}, showing judgment inconsistencies of GPT-4-as-a-judge.}
    \label{tab:qual_analysis3}
\end{table*}
\begin{table*}[!ht]
    \centering
\begin{tabular}{lrrrrrrrr|r}
\toprule
MT-Bench-EN & Wr. & Role. & Reas. & Math & Code & Extr. & STEM & Hum. & Avg. \\
\midrule
24EU-1T-pre-train              & 1.0   & 1.75  & 1.0   & 1.0   & 1.25  & 1.0   & 1.0   & 1.0   & 1.12  \\
Bactr.-DE                    & 5.1   & 5.1   & 2.0   & 1.35  & 1.95  & 2.8   & 4.25  & 5.35  & 3.49  \\
Bactr.-DEFRITES              & 4.92  & 4.0   & 2.05  & 1.05  & 1.85  & 2.75  & 5.15  & 4.15  & 3.24  \\
Bactr.-DEFRITES-sampled      & 4.4   & 4.2   & 2.45  & 1.4   & 1.65  & 2.2   & 4.8   & 4.53  & 3.2   \\
Bactr.-EN                    & 4.68  & 5.2   & 3.4   & 1.15  & 2.05  & 2.6   & 5.5   & 6.2   & 3.85  \\
Bactr.-ENDEFRITES            & 5.4   & 4.75  & 3.45  & 1.15  & 2.0   & 2.55  & 4.8   & 5.35  & 3.68  \\
Bactr.-ENDEFRITES-sampled    & 4.6   & 4.95  & 3.2   & 1.05  & 2.1   & 2.45  & 4.65  & 5.4   & 3.55  \\
Bactr.-ES                    & 5.5   & 5.1   & 2.2   & 1.35  & 2.0   & 2.6   & 4.05  & 5.48  & 3.53  \\
Bactr.-FR                    & 5.68  & 4.95  & 3.2   & 1.4   & 2.1   & 2.5   & 4.28  & 5.2   & 3.66  \\
Bactr.-IT                    & 4.82  & 4.45  & 2.75  & 1.35  & 1.89  & 2.3   & 4.35  & 4.7   & 3.33  \\
Lima-DE                        & 2.8   & 3.15  & 1.5   & 1.15  & 1.4   & 1.3   & 1.85  & 1.6   & 1.84  \\
Lima-DEFRITES                  & 2.55  & 3.65  & 2.7   & 1.35  & 1.4   & 1.65  & 3.4   & 2.05  & 2.34  \\
Lima-DEFRITES-sampled          & 2.4   & 2.45  & 3.1   & 1.1   & 1.1   & 1.05  & 1.75  & 1.25  & 1.78  \\
Lima-EN                        & 2.55  & 4.25  & 1.8   & 1.0   & 1.25  & 1.4   & 1.9   & 1.65  & 1.98  \\
Lima-ENDEFRITES                & 2.48  & 3.85  & 1.45  & 1.0   & 1.4   & 1.6   & 1.5   & 1.95  & 1.9   \\
Lima-ENDEFRITES-sampled        & 2.6   & 3.45  & 2.05  & 1.25  & 1.2   & 1.15  & 2.95  & 2.0   & 2.08  \\
Lima-ES                        & 2.8   & 3.35  & 3.55  & 1.2   & 1.35  & 1.65  & 2.35  & 1.7   & 2.24  \\
Lima-FR                        & 2.5   & 3.65  & 2.1   & 1.0   & 1.25  & 1.35  & 1.5   & 1.75  & 1.89  \\
Lima-IT                        & 2.6   & 2.3   & 2.1   & 1.1   & 1.35  & 1.1   & 1.8   & 1.7   & 1.76  \\
Bactr.-X-Avg.                & 5.01  & 4.74  & 2.74  & 1.25  & 1.95  & 2.53  & 4.65  & 5.15  & 3.5   \\
Lima-X-Avg.                    & 2.59  & 3.34  & 2.26  & 1.13  & 1.3   & 1.36  & 2.11  & 1.74  & 1.98  \\
\midrule
MT-Bench-DE & Wr. & Role. & Reas. & Math & Code & Extr. & STEM & Hum. & Avg. \\
\midrule
24EU-1T-pre-train              & 1.1   & 2.2   & 1.0   & 1.0   & 1.0   & 1.0   & 1.0   & 1.0   & 1.16  \\
Bactr.-DE                    & 5.15  & 4.6   & 3.15  & 1.4   & 1.7   & 2.1   & 4.55  & 5.7   & 3.54  \\
Bactr.-DEFRITES              & 5.48  & 4.95  & 3.1   & 1.85  & 1.8   & 2.7   & 4.65  & 6.4   & 3.87  \\
Bactr.-DEFRITES-sampled      & 4.97  & 4.15  & 2.8   & 1.5   & 1.9   & 2.45  & 4.92  & 4.58  & 3.41  \\
Bactr.-EN                    & 4.9   & 4.4   & 3.4   & 1.55  & 1.85  & 2.25  & 3.4   & 4.22  & 3.25  \\
Bactr.-ENDEFRITES            & 5.6   & 5.35  & 2.95  & 1.65  & 1.85  & 2.55  & 5.05  & 6.0   & 3.88  \\
Bactr.-ENDEFRITES-sampled    & 5.08  & 4.82  & 2.25  & 1.05  & 2.0   & 1.65  & 3.75  & 4.15  & 3.09  \\
Bactr.-ES                    & 4.75  & 4.6   & 3.5   & 1.5   & 2.05  & 2.0   & 3.6   & 4.08  & 3.26  \\
Bactr.-FR                    & 5.3   & 4.3   & 3.05  & 1.95  & 2.25  & 1.8   & 3.35  & 3.3   & 3.16  \\
Bactr.-IT                    & 4.8   & 4.55  & 2.6   & 1.5   & 2.1   & 2.6   & 3.75  & 3.65  & 3.19  \\
Lima-DE                        & 2.4   & 2.92  & 1.8   & 1.0   & 1.3   & 1.1   & 1.8   & 1.15  & 1.68  \\
Lima-DEFRITES                  & 3.5   & 3.58  & 3.2   & 1.0   & 1.1   & 1.4   & 1.7   & 2.25  & 2.22  \\
Lima-DEFRITES-sampled          & 2.0   & 2.8   & 1.4   & 1.55  & 1.0   & 1.45  & 1.2   & 1.0   & 1.55  \\
Lima-EN                        & 2.95  & 3.55  & 1.1   & 1.3   & 1.45  & 1.1   & 1.6   & 1.65  & 1.84  \\
Lima-ENDEFRITES                & 2.8   & 2.7   & 1.55  & 1.35  & 1.2   & 1.25  & 1.55  & 1.05  & 1.68  \\
Lima-ENDEFRITES-sampled        & 2.9   & 2.65  & 2.1   & 1.4   & 1.1   & 1.5   & 2.15  & 1.6   & 1.92  \\
Lima-ES                        & 2.7   & 2.45  & 1.4   & 1.35  & 1.25  & 1.3   & 2.0   & 1.6   & 1.76  \\
Lima-FR                        & 2.75  & 2.6   & 1.85  & 1.05  & 1.05  & 1.15  & 1.35  & 1.0   & 1.6   \\
Lima-IT                        & 2.88  & 3.05  & 1.4   & 1.0   & 1.05  & 1.15  & 1.5   & 1.15  & 1.65  \\
Bactr.-X-Avg.                & 5.11  & 4.64  & 2.98  & 1.55  & 1.94  & 2.23  & 4.11  & 4.68  & 3.41  \\
Lima-X-Avg.                    & 2.76  & 2.92  & 1.76  & 1.22  & 1.17  & 1.27  & 1.65  & 1.38  & 1.77  \\
\bottomrule
\end{tabular}
    \caption{Single evaluation results by GPT-4 for MT-Bench-EN and MT-Bench-DE for 24EU-7B-based models.}
    \label{tab:GPT-4-single-score-Mt-bench-ENDE}
\end{table*}

\begin{table*}[!ht]
    \centering
\begin{tabular}{lrrrrrrrr|r}
\toprule
MT-Bench-FR & Wr. & Role. & Reas. & Math & Code & Extr. & STEM & Hum. & Avg. \\
\midrule
24EU-1T-pre-train              & 0.95  & 1.1   & 0.79  & 0.68  & 0.95  & 1.0   & 0.74  & 0.75  & 0.87  \\
Bactr.-DE                    & 5.42  & 4.8   & 2.35  & 1.0   & 1.7   & 2.45  & 3.9   & 5.05  & 3.33  \\
Bactr.-DEFRITES              & 5.42  & 4.85  & 2.55  & 1.3   & 2.0   & 2.5   & 5.15  & 6.6   & 3.8   \\
Bactr.-DEFRITES-sampled      & 4.5   & 4.3   & 2.2   & 1.85  & 2.0   & 2.15  & 4.05  & 4.4   & 3.18  \\
Bactr.-EN                    & 4.25  & 4.3   & 2.9   & 1.35  & 1.65  & 2.25  & 3.9   & 3.95  & 3.07  \\
Bactr.-ENDEFRITES            & 5.15  & 4.75  & 2.6   & 1.05  & 1.75  & 2.65  & 4.95  & 4.32  & 3.4   \\
Bactr.-ENDEFRITES-sampled    & 4.15  & 3.85  & 2.45  & 1.05  & 1.85  & 3.05  & 3.9   & 3.42  & 2.97  \\
Bactr.-ES                    & 4.0   & 4.35  & 2.35  & 1.1   & 2.0   & 2.55  & 3.05  & 3.7   & 2.89  \\
Bactr.-FR                    & 4.55  & 4.9   & 2.55  & 1.65  & 1.8   & 3.0   & 4.75  & 5.65  & 3.61  \\
Bactr.-IT                    & 4.8   & 4.8   & 2.65  & 1.4   & 1.75  & 3.7   & 3.85  & 3.1   & 3.26  \\
Lima-DE                        & 1.65  & 1.55  & 1.7   & 1.0   & 1.1   & 1.1   & 1.75  & 1.6   & 1.43  \\
Lima-DEFRITES                  & 3.45  & 3.9   & 2.75  & 1.2   & 1.45  & 1.7   & 2.55  & 1.9   & 2.36  \\
Lima-DEFRITES-sampled          & 1.7   & 1.6   & 1.5   & 1.0   & 1.15  & 1.25  & 1.45  & 1.3   & 1.37  \\
Lima-EN                        & 2.4   & 3.35  & 1.7   & 1.35  & 1.1   & 1.6   & 1.7   & 1.3   & 1.81  \\
Lima-ENDEFRITES                & 2.85  & 2.4   & 1.85  & 1.2   & 1.3   & 2.2   & 1.4   & 1.4   & 1.82  \\
Lima-ENDEFRITES-sampled        & 2.25  & 2.75  & 1.8   & 1.05  & 1.4   & 1.4   & 1.85  & 1.45  & 1.74  \\
Lima-ES                        & 1.7   & 3.45  & 1.95  & 1.1   & 1.1   & 1.2   & 1.6   & 1.3   & 1.67  \\
Lima-FR                        & 2.35  & 3.25  & 2.55  & 1.0   & 1.5   & 1.45  & 2.25  & 1.35  & 1.96  \\
Lima-IT                        & 2.45  & 3.1   & 2.45  & 1.1   & 1.15  & 1.45  & 1.5   & 1.3   & 1.81  \\
Bactr.-X-Avg.                & 4.69  & 4.54  & 2.51  & 1.31  & 1.83  & 2.7   & 4.17  & 4.47  & 3.28  \\
Lima-X-Avg.                    & 2.31  & 2.82  & 2.03  & 1.11  & 1.25  & 1.48  & 1.78  & 1.43  & 1.78  \\
\midrule
MT-Bench-IT & Wr. & Role. & Reas. & Math & Code & Extr. & STEM & Hum. & Avg. \\
\midrule
24EU-1T-pre-train              & 1.0   & 2.1   & 0.93  & 1.33  & 1.0   & 1.1   & 0.85  & 0.85  & 1.15  \\
Bactr.-DE                    & 4.95  & 4.6   & 2.65  & 1.7   & 1.8   & 3.15  & 4.5   & 5.5   & 3.61  \\
Bactr.-DEFRITES              & 4.75  & 5.15  & 2.15  & 1.2   & 1.7   & 2.05  & 4.85  & 5.35  & 3.4   \\
Bactr.-DEFRITES-sampled      & 4.1   & 3.95  & 2.25  & 1.05  & 1.95  & 2.5   & 4.85  & 5.3   & 3.24  \\
Bactr.-EN                    & 4.4   & 4.5   & 2.3   & 1.2   & 1.35  & 3.1   & 4.4   & 3.75  & 3.12  \\
Bactr.-ENDEFRITES            & 4.88  & 5.5   & 2.15  & 1.4   & 2.0   & 2.65  & 6.0   & 5.22  & 3.72  \\
Bactr.-ENDEFRITES-sampled    & 4.55  & 4.9   & 2.25  & 1.0   & 1.65  & 2.3   & 4.15  & 3.75  & 3.07  \\
Bactr.-ES                    & 4.68  & 5.05  & 2.45  & 1.1   & 1.7   & 2.45  & 3.95  & 5.0   & 3.3   \\
Bactr.-FR                    & 4.58  & 4.9   & 2.15  & 1.05  & 1.3   & 2.95  & 3.75  & 4.0   & 3.08  \\
Bactr.-IT                    & 4.85  & 5.1   & 2.55  & 1.1   & 1.65  & 3.0   & 4.9   & 3.95  & 3.39  \\
Lima-DE                        & 1.85  & 2.2   & 1.1   & 1.35  & 1.15  & 1.7   & 1.1   & 1.35  & 1.48  \\
Lima-DEFRITES                  & 2.15  & 3.35  & 2.55  & 1.0   & 1.45  & 1.95  & 2.65  & 1.9   & 2.12  \\
Lima-DEFRITES-sampled          & 1.9   & 3.5   & 1.05  & 1.3   & 1.0   & 1.3   & 1.5   & 1.0   & 1.57  \\
Lima-EN                        & 2.45  & 3.15  & 1.5   & 1.1   & 1.45  & 1.4   & 1.8   & 1.2   & 1.76  \\
Lima-ENDEFRITES                & 2.45  & 2.7   & 1.7   & 1.15  & 1.35  & 1.6   & 1.5   & 1.65  & 1.76  \\
Lima-ENDEFRITES-sampled        & 1.85  & 2.9   & 2.4   & 1.0   & 1.2   & 1.6   & 1.35  & 1.5   & 1.72  \\
Lima-ES                        & 1.5   & 2.9   & 1.75  & 1.3   & 1.4   & 1.7   & 1.8   & 1.55  & 1.74  \\
Lima-FR                        & 1.85  & 2.15  & 1.35  & 1.35  & 1.25  & 1.4   & 1.1   & 1.7   & 1.52  \\
Lima-IT                        & 2.05  & 3.35  & 1.8   & 1.0   & 1.35  & 1.8   & 1.35  & 1.2   & 1.74  \\
Bactr.-X-Avg.                & 4.64  & 4.85  & 2.32  & 1.2   & 1.68  & 2.68  & 4.59  & 4.65  & 3.33  \\
Lima-X-Avg.                    & 2.01  & 2.91  & 1.69  & 1.17  & 1.29  & 1.61  & 1.57  & 1.45  & 1.71  \\
\bottomrule
\end{tabular}
    \caption{Single evaluation results by GPT-4 for MT-Bench-FR and MT-Bench-IT for 24EU-7B-based models.}
    \label{tab:GPT-4-single-score-Mt-bench-FRIT}
\end{table*}

\begin{table*}[!ht]
    \centering
\begin{tabular}{lrrrrrrrr|r}
\toprule
MT-Bench-ES & Wr. & Role. & Reas. & Math & Code & Extr. & STEM & Hum. & Avg. \\
\midrule
24EU-1T-pre-train              & 0.84  & 1.4   & 0.85  & 0.85  & 0.95  & 0.82  & 1.94  & 1.0   & 1.08  \\
Bactr.-DE                    & 4.5   & 5.18  & 2.6   & 1.05  & 1.6   & 2.95  & 4.5   & 5.6   & 3.5   \\
Bactr.-DEFRITES              & 4.47  & 5.3   & 2.1   & 1.05  & 1.85  & 1.85  & 4.55  & 4.9   & 3.26  \\
Bactr.-DEFRITES-sampled      & 4.47  & 5.18  & 1.85  & 1.4   & 1.85  & 2.45  & 4.2   & 4.0   & 3.18  \\
Bactr.-EN                    & 4.58  & 4.05  & 2.4   & 1.1   & 1.15  & 2.95  & 3.48  & 4.32  & 3.0   \\
Bactr.-ENDEFRITES            & 4.82  & 5.0   & 2.3   & 1.05  & 2.15  & 2.95  & 5.18  & 5.65  & 3.64  \\
Bactr.-ENDEFRITES-sampled    & 4.42  & 4.85  & 2.4   & 1.05  & 1.9   & 2.45  & 4.95  & 4.35  & 3.3   \\
Bactr.-ES                    & 4.15  & 4.8   & 3.0   & 1.1   & 2.1   & 2.0   & 5.05  & 5.05  & 3.41  \\
Bactr.-FR                    & 4.97  & 4.68  & 2.45  & 1.05  & 1.95  & 2.05  & 4.4   & 5.28  & 3.35  \\
Bactr.-IT                    & 4.72  & 4.15  & 2.4   & 1.15  & 1.65  & 2.1   & 4.5   & 4.25  & 3.12  \\
Lima-DE                        & 2.35  & 2.3   & 1.25  & 1.25  & 1.1   & 1.05  & 1.35  & 1.25  & 1.49  \\
Lima-DEFRITES                  & 3.2   & 3.7   & 1.3   & 1.0   & 1.0   & 1.95  & 2.3   & 2.05  & 2.06  \\
Lima-DEFRITES-sampled          & 1.8   & 2.15  & 1.4   & 1.0   & 1.0   & 1.35  & 1.9   & 1.25  & 1.48  \\
Lima-EN                        & 2.35  & 2.2   & 1.1   & 1.05  & 1.05  & 1.35  & 1.45  & 1.55  & 1.51  \\
Lima-ENDEFRITES                & 1.95  & 2.7   & 1.3   & 1.05  & 1.0   & 1.6   & 1.6   & 1.2   & 1.55  \\
Lima-ENDEFRITES-sampled        & 2.65  & 2.85  & 1.65  & 1.05  & 1.0   & 1.25  & 2.95  & 1.75  & 1.89  \\
Lima-ES                        & 2.3   & 2.6   & 1.35  & 1.0   & 1.0   & 1.15  & 1.05  & 1.4   & 1.48  \\
Lima-FR                        & 1.85  & 2.9   & 1.3   & 1.3   & 1.5   & 1.4   & 1.15  & 1.1   & 1.56  \\
Lima-IT                        & 1.6   & 2.45  & 1.2   & 1.05  & 1.0   & 1.3   & 1.05  & 1.8   & 1.43  \\
Bactr.-X-Avg.                & 4.57  & 4.8   & 2.39  & 1.11  & 1.8   & 2.42  & 4.53  & 4.82  & 3.3   \\
Lima-X-Avg.                    & 2.23  & 2.65  & 1.32  & 1.08  & 1.07  & 1.38  & 1.64  & 1.48  & 1.61  \\
\bottomrule
\end{tabular}
    \caption{Single evaluation results by GPT-4 for MT-Bench-ES for 24EU-7B-based models.}
    \label{tab:GPT-4-single-score-Mt-bench-ES}
\end{table*}

\end{document}